%% file: main.tex
\pdfoutput=1
\documentclass[10pt]{article}
\usepackage[preprint]{tmlr}
\usepackage{hyperref}
\usepackage{url}
\usepackage{amsmath,amssymb}
\makeatletter
\@ifundefined{KOMAClassName}{
  \IfFileExists{parskip.sty}{%
    \usepackage{parskip}
  }{
    \setlength{\parindent}{0pt}
    \setlength{\parskip}{6pt plus 2pt minus 1pt}}
}{
  \KOMAoptions{parskip=half}}
\makeatother
\usepackage{longtable,booktabs,array}
\usepackage{caption}
\usepackage{calc} 
\usepackage{etoolbox}
\makeatletter
\patchcmd\longtable{\par}{\if@noskipsec\mbox{}\fi\par}{}{}
\makeatother
\IfFileExists{footnotehyper.sty}{\usepackage{footnotehyper}}{\usepackage{footnote}}
\makesavenoteenv{longtable}
\usepackage{graphicx}
\makeatletter
\newsavebox\pandoc@box
\newcommand*\pandocbounded[1]{
  \sbox\pandoc@box{#1}%
  \Gscale@div\@tempa{\textheight}{\dimexpr\ht\pandoc@box+\dp\pandoc@box\relax}%
  \Gscale@div\@tempb{\linewidth}{\wd\pandoc@box}%
  \ifdim\@tempb\p@<\@tempa\p@\let\@tempa\@tempb\fi
  \ifdim\@tempa\p@<\p@\scalebox{\@tempa}{\usebox\pandoc@box}%
  \else\usebox{\pandoc@box}%
  \fi%
}
\def\fps@figure{htbp}
\makeatother
\providecommand{\tightlist}{%
  \setlength{\itemsep}{0pt}\setlength{\parskip}{0pt}}
\usepackage[]{natbib}
\input{layout.tex}
\hypersetup{pdftitle={Where Generated Floor Plans Stop: Calibrated Sample-Level Checks Against a Built Baseline}, pdfauthor={Yiliang
Shao}, hidelinks}

\title{Where Generated Floor Plans Stop: Calibrated Sample-Level Checks
Against a Built Baseline}
\author{\name Yiliang Shao \email yiliangsh@gmail.com \\ \addr Skidmore,
Owings \& Merrill (SOM), Chicago}

\begin{document}

\maketitle

\begin{abstract}
Floor plan generators are judged mainly by Compatibility, a graph edit
distance to the requested adjacencies that is near saturation and blind
to room shape. Our sample-level protocol calibrates each threshold no
standard states at a declared false-rejection rate on real samples and
grades them with the same checks. It reports each sample at its first
failure, counts samples without a verdict apart, and attributes a stop
to the generator or to its representation where a control exists. We
apply it, with a production layout solver's checks, to four published
generators against two built corpora, RPLAN and Modified Swiss Dwellings
(MSD). A squareness check set per room type refuses 5 per cent of RPLAN
apartments and 33.7 per cent of House-GAN++ plans. The gap survives
resolution, tracing and orientation controls. Pooled over rooms, a
Wasserstein distance on squareness orders the generators nearly as the
check does but separates House-GAN++ from the built corpora by far less.
Taken per plan on the check's worst-room margin, it separates them about
fivefold to sevenfold. Within the GAN generators, Compatibility barely
predicts which plans pass. We release the checker, which reproduces the
solver's first failing check on 3,998 held-out RPLAN plans.
\end{abstract}

\section{1. Introduction}\label{introduction}

Floor plan generators are judged mainly by Compatibility, the graph edit
distance between the adjacency graph a generator was asked for and the
one recovered from its output. The measure is cheap, automatic,
comparable across model families and defined without reference to any
building. On RPLAN it is approaching saturation. For example,
\citet{lara2026rlvr} report Compatibility of 0.01 on five-room RPLAN
tasks and 0.15 on eight-room tasks, against 1.5 and 2.5 for the
strongest prior diffusion baseline. They concede in their limitations
that their rewards ``are necessary but not sufficient for architectural
validity''.

The common criteria for evaluating generative models are largely
independent of one another, and a model should be judged against its
application \citep{theis2016note}. Some applications demand that each
output satisfy hard constraints. There, a score that compares samples
with a request or with a reference distribution can rise while
individual samples remain invalid. In molecular docking, per-pose
validity checks showed that methods claiming state-of-the-art accuracy
on the standard metric often produce physically implausible poses
\citep{buttenschoen2024posebusters}. Floor plan generation is a further
case, except that its shape constraints have no physical constant and no
building code states them. Their thresholds must therefore first be
calibrated. So the checks here test shape conventions calibrated on
built plans, not validity. By a false rejection we mean only a built
plan that a check refuses. Its rate is set by convention, and we do not
test whether such plans would be accepted in practice.

The alternatives leave the same gap. FID \citep{heusel2017fid} measures
resemblance to the training corpus, and human preference studies and
learned preference scores record what people prefer (Section 2). None of
them tests each plan against explicit checks or names the rule a plan
breaks. Automated rule checking of building models returns a located,
named violation \citep{eastman2009rulechecking}. The model codes it
draws on, however, state no shape threshold for rooms. The rule-based
evaluations of generated plans include GreenPlanner
\citep{zeng2025greenplanner} and \citet{park2024quality}. They neither
calibrate a shape threshold on built plans nor grade built plans by the
same rules as a baseline (Section 2).

We therefore propose a protocol for sample-level evaluation against a
built baseline. Every threshold that no standard states is calibrated on
real samples at a declared rate of false rejection. The real samples are
graded with the same checks, so that the built baseline fails at a known
rate rather than being assumed to pass. Each sample is reported at the
first check it fails. Samples the checks cannot judge are counted apart
from failures. Each stop is attributed either to the generator or to the
way it represents a plan, by grading the real samples in that
representation, where such a rendering exists; otherwise the attribution
is left open. This study asks three research questions:

\begin{itemize}
\tightlist
\item
  \textbf{RQ1.} Can the checks of a production layout solver be carried
  to public floor plan data, with each threshold no standard states
  calibrated so that built plans fail at a declared rate?
\item
  \textbf{RQ2.} At which checks do the plans of published generators
  fail, and how do they compare with built plans?
\item
  \textbf{RQ3.} Does Compatibility predict whether a generated plan
  passes these checks?
\end{itemize}

We demonstrate the protocol with six checks adapted from the acceptance
ladder of a production layout solver. The ladder is the ordered sequence
of checks after which the solver accepts or rejects a plan. Three of the
checks run on public data. The first is room outlines, which rejects no
plan. The other two compare room shapes with thresholds fitted from real
plans. Minimum compactness relates a room's area to that of a circle
with the same perimeter. Minimum squareness, set for each room type, is
the ratio of short side to long. Each threshold is fitted so that its
check alone refuses 5 per cent of real built plans. Both stand in for
whether a room can be furnished, which this paper does not test.

Four published generators are audited against the two built corpora,
RPLAN and the Modified Swiss Dwellings (MSD) apartments. The ladder can
stop a plan without judging it, so rates are quoted over three
denominators, written G, A and V throughout. The graded plans G are
every plan a generator returned and every built plan sampled. The
admitted plans A leave out the plans refused at ingest, the plans that
cannot be read and the plans that fail to build. The plans V that
reached a verdict also leave out the plans halted without a verdict at a
precondition of the fourth check. This precondition requires the walls
of each room to meet at right angles. The paper makes three
contributions.

\begin{itemize}
\tightlist
\item
  \textbf{C1, the protocol with its transfer specification (RQ1).} The
  protocol as set out above, each check with a named rejection reason,
  rates over G, A and V and the shape thresholds fitted at 5 per cent. A
  transfer specification records which of the solver's checks public
  corpora can carry and where each threshold comes from. What transfers
  is the form of the checks, not their values: no production threshold
  value is used. From the protocol follow four quantities a generator
  paper can report beside Compatibility (Section 5.1).
\item
  \textbf{C2, a controlled shape result for House-GAN++ (RQ2, RQ3).}
  HouseDiffusion stops at ingest, GSDiff at the right-angle precondition
  and House-GAN++ at the proportion check (Section 4.2.1). Set per room
  type, the proportion check alone refuses 33.7 per cent of the
  House-GAN++ plans the checker admits, against 5.0 per cent of RPLAN
  and 4.9 per cent of the MSD apartments. Controls for resolution,
  tracing and orientation move that separation by about a point or less
  with the thresholds unchanged (Section 4.2.2). House-GAN is also
  refused by the proportion check read alone. RPLAN drawn as House-GAN's
  training targets is refused as often, however, so that refusal belongs
  to its representation. The whole-ladder margin of the MSD apartments
  over House-GAN++ is 12.1, 5.5 and 23.5 points over G, A and V. It
  decomposes into the proportion check, which carries it, and two
  representation terms, ingest refusal and the right-angle precondition.
  No exchanged fit inverts it, although over admitted plans on the other
  corpus's fit the interval spans zero, -0.8 to +4.1, when whole
  buildings are resampled. At the 5 per cent rate it is positive in
  every held-out split over G and V and on the mean of the splits over A
  (Section 4.2.3). Within House-GAN and House-GAN++, Compatibility as
  recovered from the plans barely predicts a pass. Its area under the
  ROC curve lies within 0.07 of chance in either direction (Section
  5.1). Most diffusion output stops at ingest or at the precondition, so
  the protocol does not grade it and does not settle whether these stops
  belong to the generator or to its representation.
\item
  \textbf{C3, a released checker validated on held-out plans.} The
  solver runs privately as a reference implementation, or oracle. On
  3,998 held-out RPLAN plans the public checker names the same first
  failing check as the solver for 100.0 per cent, including all 327
  rejections. On the audited plans they agree for 99.6 per cent of the
  plans both admit and 99.8 per cent of those both return a verdict on.
  The checker was corrected against these plans, however, so that
  agreement is not independent evidence (Section 4.1). The ingest and
  right-angle precondition paths, which carry the margins over G and A,
  have no held-out test, since RPLAN produces no ingest refusal and no
  precondition stop.
\end{itemize}

\section{2. Related work}\label{related-work}

Learning-based floor plan generation rests on one dataset and one
interface. The dataset is RPLAN \citep{wu2019rplan}, 80,788
single-apartment plans from real residential buildings distributed on
request. The interface, a layout graph in and a plan out, was fixed by
Graph2Plan \citep{hu2020graph2plan}. The generators audited here differ
chiefly in how they represent a plan. House-GAN
\citep{nauata2020housegan} and House-GAN++ \citep{nauata2021housegan2}
generate one raster mask per room. The audit traces each mask into
axis-aligned rectilinear polygons. HouseDiffusion
\citep{shabani2023housediffusion} emits polygon vertices, and GSDiff
\citep{hu2025gsdiff} adds a wall-junction graph and an alignment loss.
Other recent work varies the task, the model family or the conditioning
rather than the evaluation, almost all of it on RPLAN
\citep{yin2025floorplanmae, yin2025fpds, klimenko2026hypergraphformer, lara2026rlvr, luo2022floorplangan, aalaei2023graphgan, zeng2024multiconditional, shekhawat2021dimensioned}.
The one exception is the MSD benchmark \citep{vanengelenburg2024msd},
5,300 multi-unit plans covering 18,900 apartments, with a HouseDiffusion
adaptation \citep{kuhn2023msdhousediffusion}.

Compatibility is the graph edit distance \citep{sanfeliu1983distance}
between the input graph and the one recovered from the plan. The metrics
of DStruct2Design \citep{luo2024dstruct2design} likewise measure
conformance to the request, and SSIG \citep{vanengelenburg2023ssig} adds
image overlap. The dual-aspect framework of \citet{guo2025dualaspect},
applied to 240 pix2pix models, scores pixel similarity to reference
plans and the continuity of segmentation lines. These measures compare a
plan with another plan, not with a rule. FID \citep{heusel2017fid}
certifies typicality rather than correctness, and human preference
studies \citep{nauata2021housegan2} cannot be rerun by the next author.
Measures that moved beyond a single score \citep{theis2016note}, such as
precision and recall
\citep{sajjadi2018precision, kynkaanniemi2019improved} and density and
coverage \citep{naeem2020reliable}, still compare with a reference
sample whose defects remain the target. Molecular
\citep{brown2019guacamol, polykovskiy2020moses} and layout
\citep{kikuchi2021layoutganpp, inoue2023layoutdm} benchmarks report
validity as a fraction of the sample or a continuous score. Rule-based
evaluations come closer. Among them, \citet{park2024quality} test
relational GAN layouts against rules of thumb on room size, privacy and
connectivity. GreenPlanner \citep{zeng2025greenplanner} scores
generators with an evaluator trained on RPLAN plans labelled for energy
use and Chinese code rules. It finds 37.4 per cent of RPLAN plans fully
compliant and reads the rest as defects to filter from the training
data. RFP-A \citep{zeng2025rfpa} proposes four rule-based measures.
FloorPlan-LLaMa \citep{yin2025floorplanllama} learns from architect
ratings a preference score that cannot name the rule a plan broke. In
\citet{bu2027layoutquality}, plans are scored against experience-based
rules retrieved by a multimodal language model, validated against expert
preferences. Space syntax
\citep{hillier1984sociallogic, turner2001isovists} describes how a plan
performs, and the references \citet{jiang2026spacesyntax} derive from
real dwellings measure configuration rather than room shape. Three
surveys cover the field
\citep{weber2022review, jang2025generative, liu2025layoutreview}, and
\citet{mostafavi2025interplay} argue the gap on the design side.

The closest analogue is PoseBusters \citep{buttenschoen2024posebusters},
which applies chemical and physical validity checks to each pose a
docking method generates and finds that methods claiming
state-of-the-art accuracy often produce physically implausible
structures. The protocol shares that design but orders its checks, so
that failures are located rather than counted. It also calibrates its
shape thresholds on built plans at a declared rate, where PoseBusters
takes physical constants.

Rule checking of designs
\citep{eastman2009rulechecking, nawari2019gaf, chen2024arceak, gautam2026compliance},
with which the ladder belongs, returns a located, named violation rather
than a distance and is largely disconnected from generation. Its rules
come from model codes, the International Building Code
\citep{icc2024ibc}, the International Residential Code
\citep{icc2021irc}, the ADA Standards \citep{doj2010ada} and ICC A117.1
\citep{icc2017a117}, and none states a room compactness threshold or
aspect limit. Their one shape constraint, the IRC's least horizontal
dimension of 7 feet, is tested by the solver during construction and
measured on finished plans in Section 4.2.3.

What is new here lies in the protocol and its baseline. The metrics
above return a score against a request or a reference sample. The ladder
instead returns, for each plan, a verdict and the first check failed.
The rule-based evaluations take rules of thumb, code thresholds that
test no room shape, expert preferences or reference plans. The ladder
instead takes its checks and their order from a production layout
solver. It calibrates its shape thresholds on built stock at a declared
rate of false rejection, and so supplies the thresholds the model codes
do not state. It grades two built corpora with the same checks as four
generators. These corpora serve as a baseline rather than, as in
GreenPlanner, as data to filter. The transfer specification records
which checks a dataset can carry and the source of each threshold.

\section{3. Methods}\label{methods}

\subsection{3.1 The ladder as it is actually
graded}\label{the-ladder-as-it-is-actually-graded}

We transcribed the ladder check by check from the production solver's
source at a pinned commit into the transfer specification. The public
checker was then implemented from the specification alone, without
reference to the source (Section 3.3). The six checks run in a fixed
order, each assuming what the one above it proved. A check that rejects
returns immediately, so the checks below a plan's first failure never
run. A histogram over stages therefore states where a corpus stops
rather than everything that is wrong with it. Figure 1 draws the ladder
and Table 1 lists its checks.

\begin{figure}
\centering
\includegraphics[width=0.75\linewidth,height=\textheight,keepaspectratio,alt={The acceptance ladder as graded on public datasets. Ingest refuses plans whose rooms do not tile the outline; six checks follow in fixed order. Compactness and proportions are corpus-fitted; room appendices runs only its right-angle precondition; suite container and final adjacency do not transfer.}]{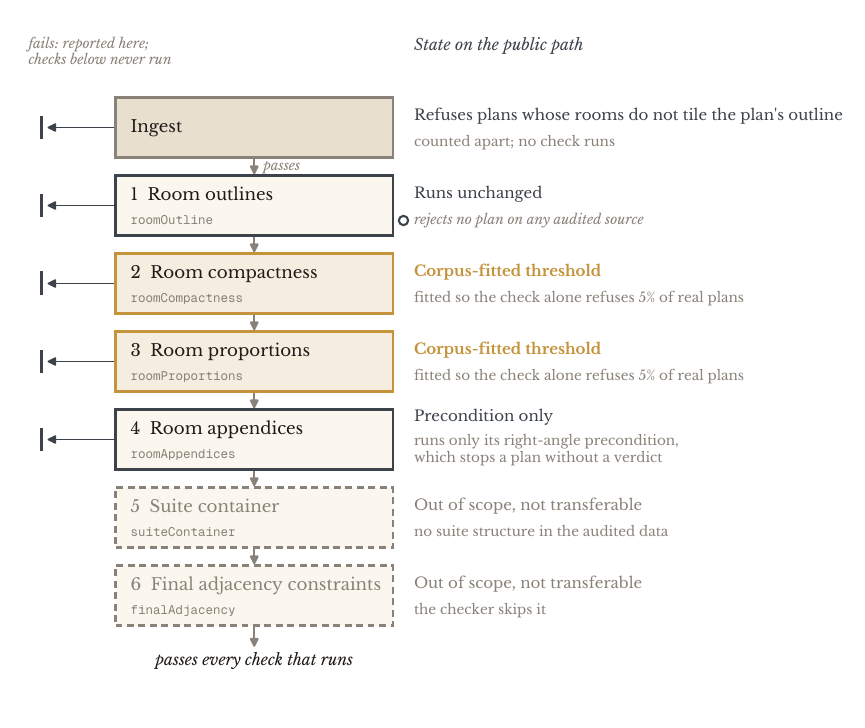}
\caption{The acceptance ladder as graded on public datasets. Ingest
refuses plans whose rooms do not tile the outline; six checks follow in
fixed order. Compactness and proportions are corpus-fitted; room
appendices runs only its right-angle precondition; suite container and
final adjacency do not transfer.}\label{fig:ladder_schematic}
\end{figure}

\begin{longtable}[]{@{}
  >{\raggedright\arraybackslash}p{(\linewidth - 6\tabcolsep) * \real{0.2178}}
  >{\raggedright\arraybackslash}p{(\linewidth - 6\tabcolsep) * \real{0.3963}}
  >{\raggedright\arraybackslash}p{(\linewidth - 6\tabcolsep) * \real{0.2457}}
  >{\raggedright\arraybackslash}p{(\linewidth - 6\tabcolsep) * \real{0.1402}}@{}}
\caption{The six checks of the solver's acceptance ladder, in the order
they run, with what each decides, where its threshold comes from and its
state on public datasets.}\tabularnewline
\toprule\noalign{}
\begin{minipage}[b]{\linewidth}\raggedright
Check
\end{minipage} & \begin{minipage}[b]{\linewidth}\raggedright
What it decides
\end{minipage} & \begin{minipage}[b]{\linewidth}\raggedright
Where the threshold comes from
\end{minipage} & \begin{minipage}[b]{\linewidth}\raggedright
On public datasets
\end{minipage} \\
\midrule\noalign{}
\endfirsthead
\toprule\noalign{}
\begin{minipage}[b]{\linewidth}\raggedright
Check
\end{minipage} & \begin{minipage}[b]{\linewidth}\raggedright
What it decides
\end{minipage} & \begin{minipage}[b]{\linewidth}\raggedright
Where the threshold comes from
\end{minipage} & \begin{minipage}[b]{\linewidth}\raggedright
On public datasets
\end{minipage} \\
\midrule\noalign{}
\endhead
\bottomrule\noalign{}
\endlastfoot
1 Room outlines \texttt{roomOutline} & Geometry alone, on a plan ingest
has accepted & No request-wide constant and no per-room requirement &
Runs; no decision recorded \\
2 Room compactness \texttt{roomCompactness} & The least compact
non-hallway room against a Polsby-Popper threshold
\citep{polsby1991compactness} & Fitted so the check alone refuses 5 per
cent of real plans of each kind: 0.365 RPLAN apartments, 0.3237 MSD
apartments, 0.2406 MSD floors; no external validation & Runs,
corpus-fitted \\
3 Room proportions \texttt{roomProportions} & Each room's squareness,
short side over long of its unrotated bounding box, against a threshold
per room type & One quantile of each room type's squareness, chosen so
the check alone refuses 5 per cent of real plans of each kind; no
external validation & Runs, corpus-fitted \\
4 Room appendices \texttt{roomAppendices} & Whether a protruding
sub-area counts as usable space, once the room's walls form a closed
loop meeting at right angles & One of the two conditions in each of the
policy's two tests derived; the precondition's tolerance derived &
Policy disabled; precondition stops the plan without a verdict \\
5 Suite container \texttt{suiteContainer} & Whether a suite contains the
rooms it declares & None needed: the data carry no suites & Not
transferable \\
6 Final adjacency constraints \texttt{finalAdjacency} & Whether the
final geometry still meets the requested adjacency program; a request
with no program leaves it nothing to test & None needed: the data state
no adjacency program & Not transferable; reported as skipped \\
\end{longtable}

Neither corpus-fitted threshold has a public derivation. Practitioners
set the private production compactness value by their judgment of which
room shapes fall out of convention. We use no production value of either
threshold (Appendix A). The one number carried over from the solver's
proportion check is its cap on a bedroom's threshold, 0.60, which no
fitted threshold reaches.

The thresholds are fitted at the level of plans rather than rooms. Each
check judges a plan by its worst room. A room-level threshold set at the
fifth percentile of rooms would therefore alone refuse about 30 per cent
of real plans. We fit RPLAN apartments, MSD apartments and whole MSD
floors separately, so that each check alone refuses 5 per cent of the
admitted plans of each kind. A floor here is one storey of a
multi-apartment building, and a whole floor that states one apartment
counts as an apartment. For compactness the solver's grading comes
within half a point of that rate (Table 2). Generated plans are graded
on the RPLAN thresholds. The 5 per cent rate is a convention rather than
a derivation, and every margin is conditional on it (Section 4.2.3).
Unless stated otherwise, every number comes from one grading at one
commit, called the reported run. GSDiff's plans are graded at the same
thresholds by the same code in a separate pass.

A plan without a verdict was stopped rather than rejected, and Table 2
counts such plans apart, in three columns. A stop at the right-angle
precondition is counted under its check. Ingest refusal counts plans
whose rooms do not tile the outline. Other gathers plans the converter
or the ingest cannot read, whole MSD floors that divide into more than
one part, and plans the solver fails to build or loses to an error. The
named rejection reasons do not cover circulation, corridor topology, a
breach of the boundary between dwellings or door clear width. We report
no results for them.

One MSD apartment, floor 1157, apartment u5, shows the ladder at work
(Figure 5, panel b). It passes ingest, room outlines and compactness.
Its kitchen, however, is the only room in the plan below its threshold.
The kitchen's unrotated bounding box of 1.30 by 4.57 m gives a
squareness of 0.285 against the Kitchen threshold of 0.3043. The
proportion check therefore rejects the plan, and checks 4 to 6 never
run. A built kitchen 1.3 m wide is thus refused. This is the kind of
false rejection the thresholds allow on 5 per cent of built plans, a
rate set by convention without a test of whether such plans would be
accepted in practice.

\subsection{3.2 Transfer to public
datasets}\label{transfer-to-public-datasets}

The specification records, for each check and dataset, whether the check
crosses unchanged, crosses with a re-derived constant or a fitted
threshold, or does not transfer (Table 1, last column). A check with
nothing to test is reported as not transferable, since counting it as a
pass would inflate every pass rate. Check 4 transfers only as far as its
right-angle precondition. Its policy makes two tests of two conditions
each, and a public derivation exists for only one condition in each. The
suite container needs suites, but the audited plans are single
apartments or floors. Final adjacency re-tests a requested adjacency
program, which no built plan states. Sending a generated plan's
conditioning graph in its place would restate Compatibility as a pass or
a fail. A built plan's own graph would pass by construction.

Four features of the solver are not graded. Two are room exposure to
corridors and other dwellings and the area program, which never decides
a finished plan's outcome. The other two are a minimum room width and a
door clearance, which the solver reports as observations.

\begin{figure}
\centering
\includegraphics[width=0.75\linewidth,height=\textheight,keepaspectratio,alt={The proportion threshold for each room type, fitted at the plan level from each built corpus for the reported run. Squareness is the short side over the long side of a room's bounding box. Circles: RPLAN apartments. Squares: MSD apartments.}]{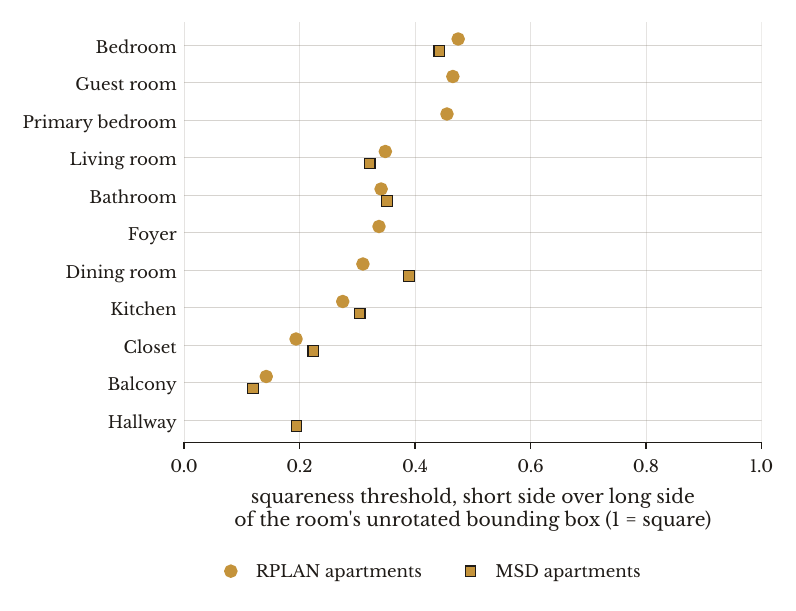}
\caption{The proportion threshold for each room type, fitted at the plan
level from each built corpus for the reported run. Squareness is the
short side over the long side of a room's bounding box. Circles: RPLAN
apartments. Squares: MSD
apartments.}\label{fig:proportion_floor_by_category}
\end{figure}

Both shape thresholds come from the corpora they grade (Figure 2), and
Section 4.2.3 measures what that contributes. Three decisions recorded
in Appendix B condition every number in Section 4.2. The first concerns
the rectify threshold, which sets how far a wall may deviate from an
axis and still be snapped straight. We set it to zero because the
publicly derived value of one RPLAN raster cell left most House-GAN++
plans unbuilt (Appendix B, decision 5). The second decision refuses at
ingest any plan whose rooms do not tile the outline. The third excludes
the MSD floors from every margin. A floor is judged by its worst room,
and it holds 30.57 room polygons on average, corridors, balconies and
closets included. An MSD apartment holds 7.86 and a House-GAN++ plan
6.27.

\subsection{3.3 The checker}\label{the-checker}

The transferable subset is released as \texttt{floorcheck}, a
one-command checker that reads a documented neutral plan schema. It
emits a verdict for each check in the named rejection reasons of Section
3.1. Its \texttt{-\/-requirements} option selects either thresholds that
need no corpus or the corpus-fitted ones. The \texttt{standards} setting
applies building-code minima and no proportion threshold. No source sets
a compactness threshold, so this setting places compactness at 0.589,
the rounded score of a three-to-one rectangle. The \texttt{dataset-p05}
setting, the default, applies the corpus-fitted thresholds and is the
source of every margin in Section 4.2. An exporter restates RPLAN, MSD
and generator output in the schema, one document per plan. Checks 1 to 3
are implemented in full, check 4 only as far as its right-angle
precondition and checks 5 and 6 not at all. The checker reports what it
declines, including the appendix policy and the area program, rather
than passing it. It is a clean-room implementation, written from the
specification alone without sight of the solver's source. It cites a
specification section for every constant. Section 4.1 validates it
against the solver, run privately as the oracle. Unless stated
otherwise, an interval on a difference of pass rates is a percentile
interval from a parametric bootstrap of the two rates.

The version tagged for this paper is floorcheck v0.1.1,
https://github.com/yiliangs/floorcheck, archived at
https://doi.org/10.5281/zenodo.23129970 \citep{shao2026floorcheck}. Its
contents, its licences and the reduced form of the deposited plans are
given under Data availability, together with the tools for the four
quantities of Section 5.1. The specification accompanies the paper as
Supplementary material 1.

\section{4. Results}\label{results}

We ran the ladder over four published generators and the ground-truth
corpora, counting every sampled plan. Two further candidates are not
audited. The RLVR generator of \citet{lara2026rlvr}, trained by
reinforcement learning with verifiable rewards, has 70-billion-parameter
checkpoints and no released sample set we could find (Supplementary
analyses, Section S3). The MSD adaptation of HouseDiffusion
\citep{kuhn2023msdhousediffusion} releases no checkpoint. RPLAN is
sampled as every twentieth of its first 80,000 plans, 4,000 of its
80,788. MSD is sampled as the 4,167 floors of its train split, with the
apartments of the first 4,000. House-GAN, House-GAN++ and HouseDiffusion
each run once on 4,000 RPLAN conditioning graphs. The graphs are drawn
at an even stride, 1,000 in each of four buckets of five to eight
requested rooms. House-GAN++ drops at least one room from 581 of its
plans. HouseDiffusion loses nine plans whose room and door corners
exceed the 100 points its model accepts. GSDiff is conditioned on its
own test split, matched to the same buckets. It yields 2,934 plans, and
2,911 of them survive its reconstruction. The solver's grading needs the
private solver, but the public checker reruns it (Section 3.3;
Supplementary analyses, Section S1, for MSD).

\subsection{4.1 Validation of the public
checker}\label{validation-of-the-public-checker}

Two implementations agree on a plan when both pass it or both reject it
at the same first failing check. Both stopping it at the right-angle
precondition also counts as agreement, but a stop on one side only
counts as a disagreement. Agreement is scoped to finished plans
(Appendix B, decision 10). On 4,000 RPLAN plans held out from the
audited sample and never corrected against, the public checker and the
solver agree on 100.0 per cent of the 3,998 both admitted. These include
all 327 plans the ladder rejects, 191 at compactness and 136 at
proportions. The test therefore covers the attribution of the first
failure and not only the pass rate.

On the reported run's plans that both ingests accept and the solver
builds, the two agree on the first failing check for 99.6 per cent (95
per cent Wilson interval \citep{wilson1927interval} 99.5 to 99.7). Over
the plans both carry to a verdict, they agree for 99.8 per cent (99.7 to
99.8, Figure 3). Supplementary analyses, Section S2, accounts for these
populations, by source and for GSDiff, and triages every disagreement.

This agreement on the audited plans shows a faithful reimplementation of
the transcription, against an oracle readers cannot inspect. It is not
independent evidence, however. Earlier gradings corrected the
specification and the checker wherever they disagreed with the solver on
these same plans (Supplementary analyses, Section S2). The independent
evidence is the held-out RPLAN test above. No such population exists for
MSD or the generators, so the ingest and right-angle precondition paths,
which carry the margins over G and A, have no held-out test.

\begin{figure}
\centering
\includegraphics[width=1\linewidth,height=\textheight,keepaspectratio,alt={Agreement between the public checker and the solver on the first failing check, per source, with 95 per cent Wilson intervals. Dashed line: the pooled 99.8 over 26,126 plans. Hollow marker: 4,000 held-out RPLAN plans.}]{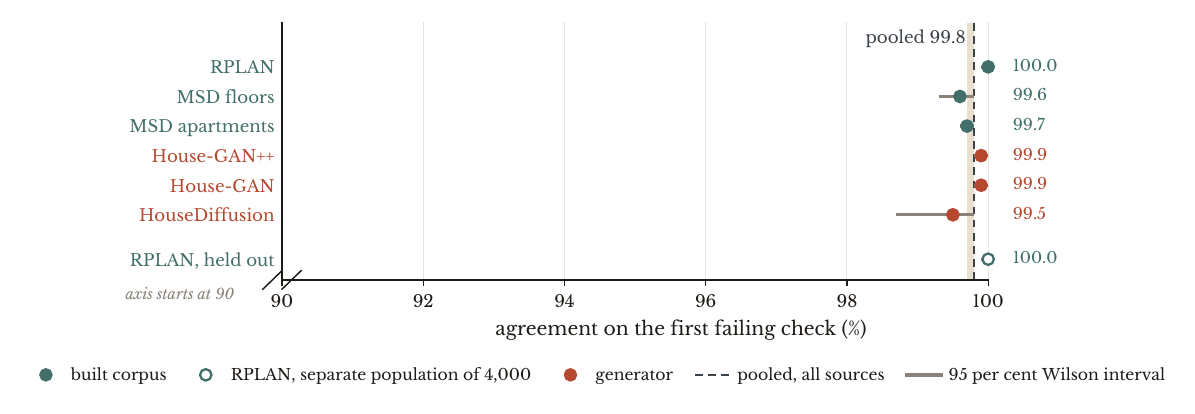}
\caption{Agreement between the public checker and the solver on the
first failing check, per source, with 95 per cent Wilson intervals.
Dashed line: the pooled 99.8 over 26,126 plans. Hollow marker: 4,000
held-out RPLAN plans.}\label{fig:checker_agreement}
\end{figure}

The solver is proprietary, and its cells cannot be rerun publicly. The
claims are therefore also stated on the public checker's own verdicts.
These give the MSD apartments a margin of +12.2 points over House-GAN++
over graded plans (interval 10.5 to 14.0), against the solver's +12.1
points. House-GAN++ released no sample set, and its regeneration is
nondeterministic on CUDA. A second set passes 2,277 of 4,000 plans
against the first set's 2,325. This moves the checker's margin to +13.4
points (11.7 to 15.2), and each estimate lies inside the other's
interval.

\subsection{4.2 Audit of published
generators}\label{audit-of-published-generators}

\subsubsection{4.2.1 Where each corpus
stops}\label{where-each-corpus-stops}

Table 2 gives, for each source, the share of graded plans that pass and
the share that stops at each check. GSDiff's shares come from its
separate pass (Section 3.1). A check that rejects returns immediately,
so the shares are mutually exclusive. Figure 4 draws the same partition.

\LTkeeptrue

\begin{longtable}[]{@{}
  >{\raggedright\arraybackslash}p{(\linewidth - 14\tabcolsep) * \real{0.2353}}
  >{\raggedright\arraybackslash}p{(\linewidth - 14\tabcolsep) * \real{0.0824}}
  >{\raggedright\arraybackslash}p{(\linewidth - 14\tabcolsep) * \real{0.0824}}
  >{\raggedright\arraybackslash}p{(\linewidth - 14\tabcolsep) * \real{0.1412}}
  >{\raggedright\arraybackslash}p{(\linewidth - 14\tabcolsep) * \real{0.1412}}
  >{\raggedright\arraybackslash}p{(\linewidth - 14\tabcolsep) * \real{0.1529}}
  >{\raggedright\arraybackslash}p{(\linewidth - 14\tabcolsep) * \real{0.0941}}
  >{\raggedright\arraybackslash}p{(\linewidth - 14\tabcolsep) * \real{0.0706}}@{}}
\caption{Where each corpus stops on the reported run, in per cent of the
plans graded for each source; a plan is counted at the first check that
stops it. The right-angle precondition is that of check 4, and it,
Ingest refusal and Other are the stops without a verdict of Section
3.1.}\tabularnewline
\toprule\noalign{}
\begin{minipage}[b]{\linewidth}\raggedright
Source
\end{minipage} & \begin{minipage}[b]{\linewidth}\raggedright
Graded
\end{minipage} & \begin{minipage}[b]{\linewidth}\raggedright
Passed
\end{minipage} & \begin{minipage}[b]{\linewidth}\raggedright
Compactness
\end{minipage} & \begin{minipage}[b]{\linewidth}\raggedright
Proportions
\end{minipage} & \begin{minipage}[b]{\linewidth}\raggedright
Right-angle precondition
\end{minipage} & \begin{minipage}[b]{\linewidth}\raggedright
Ingest refusal
\end{minipage} & \begin{minipage}[b]{\linewidth}\raggedright
Other
\end{minipage} \\
\midrule\noalign{}
\endfirsthead
\toprule\noalign{}
\begin{minipage}[b]{\linewidth}\raggedright
Source
\end{minipage} & \begin{minipage}[b]{\linewidth}\raggedright
Graded
\end{minipage} & \begin{minipage}[b]{\linewidth}\raggedright
Passed
\end{minipage} & \begin{minipage}[b]{\linewidth}\raggedright
Compactness
\end{minipage} & \begin{minipage}[b]{\linewidth}\raggedright
Proportions
\end{minipage} & \begin{minipage}[b]{\linewidth}\raggedright
Right-angle precondition
\end{minipage} & \begin{minipage}[b]{\linewidth}\raggedright
Ingest refusal
\end{minipage} & \begin{minipage}[b]{\linewidth}\raggedright
Other
\end{minipage} \\
\midrule\noalign{}
\endhead
\bottomrule\noalign{}
\endlastfoot
RPLAN ground truth & 4,000 & 91.3 & 5.0 & 3.7 & 0.0 & 0.0 & 0.1 \\
MSD ground truth, per apartment & 14,324 & 70.3 & 4.8 & 3.9 & 20.1 & 0.8
& 0.0 \\
MSD ground truth, per floor & 4,167 & 57.8 & 5.4 & 3.7 & 28.8 & 0.9 &
3.5 \\
House-GAN++ & 4,000 & 58.2 & 2.5 & 28.3 & 0.0 & 11.1 & 0.0 \\
House-GAN & 4,000 & 48.7 & 20.6 & 19.6 & 0.0 & 11.1 & 0.0 \\
HouseDiffusion & 3,991 & 1.1 & 21.1 & 1.1 & 2.4 & 70.4 & 3.9 \\
GSDiff & 2,911 & 4.0 & 3.6 & 5.3 & 83.3 & 3.1 & 0.7 \\
\end{longtable}

\LTkeepfalse

The MSD floors row is reported for coverage only (Section 3.2). Room
outlines has no column because no plan from any source first fails
there. The precondition column fills on MSD because the exact geometry
hands the solver the dataset's own wall directions. House-GAN++, by
contrast, passes the precondition by construction of its axis-aligned
tracing. The difference is a property of the representation, not a
ranking. Between the built corpora, the MSD apartments' pass rate (95
per cent Wilson interval 69.6 to 71.1) does not overlap RPLAN's (90.4 to
92.1).

\begin{figure}
\centering
\includegraphics[width=1\linewidth,height=\textheight,keepaspectratio,alt={Where each corpus stops on the reported run, for RPLAN, the MSD apartments and the four generators: passed, refused at ingest, stopped at the right-angle precondition, rejected at compactness or proportions, and other refusal or build failure. Built corpora in teal, generators in terracotta.}]{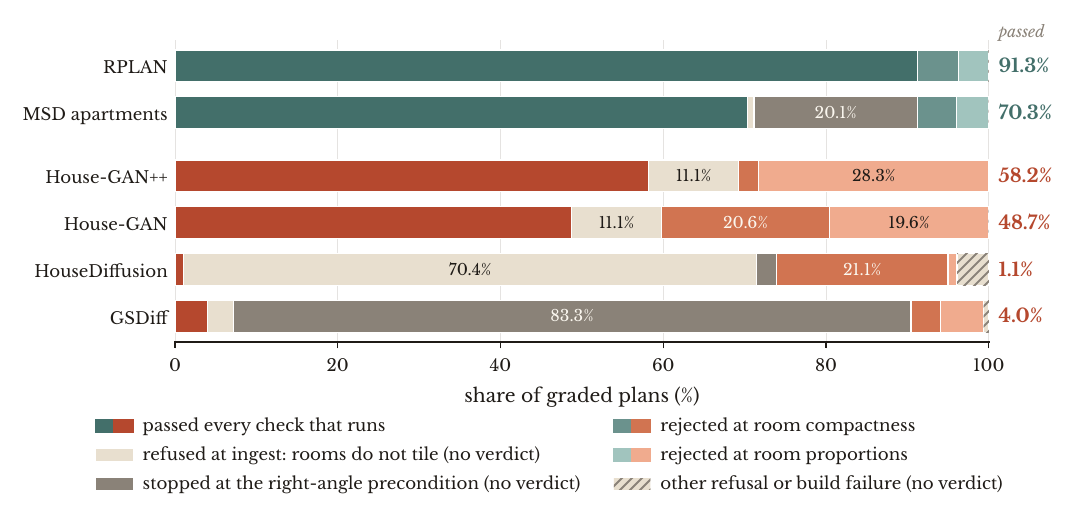}
\caption{Where each corpus stops on the reported run, for RPLAN, the MSD
apartments and the four generators: passed, refused at ingest, stopped
at the right-angle precondition, rejected at compactness or proportions,
and other refusal or build failure. Built corpora in teal, generators in
terracotta.}\label{fig:where_each_corpus_stops}
\end{figure}

A stop says something about a generator only when it is not a property
of the generator's representation. For each generator, Table 3 sets out
which of the two its main stop is and what decides it. For the two
diffusion generators no control decides it, and the attribution is left
open. Figure 5 draws one plan from each source at the check where it
stops. The overlapping rooms that Table 3 records for HouseDiffusion are
not rounding residue (Supplementary analyses, Section S3). The solver
admits only 1,024 of HouseDiffusion's 3,991 plans and passes 4.1 per
cent of those, so HouseDiffusion binds no margin. The two GANs are each
refused at ingest on 11.1 per cent of the same 4,000 graphs, but on
nearly independent sets. Both are refused on 46 graphs, against 49
expected by chance. The refusals therefore belong to each generator, not
to the graph. Nor does exposure to the training set explain the stops.
Refusal rates are alike on graphs a generator saw in training and on
graphs it did not (Supplementary analyses, Section S3).

\begin{longtable}[]{@{}
  >{\raggedright\arraybackslash}p{(\linewidth - 6\tabcolsep) * \real{0.1765}}
  >{\raggedright\arraybackslash}p{(\linewidth - 6\tabcolsep) * \real{0.1748}}
  >{\raggedright\arraybackslash}p{(\linewidth - 6\tabcolsep) * \real{0.2948}}
  >{\raggedright\arraybackslash}p{(\linewidth - 6\tabcolsep) * \real{0.3539}}@{}}
\caption{Where each generator stops on the reported run, whether the
stop is attributed to the generator or to its representation, and the
control or record that decides it; the controls are set out in Section
4.2.2.}\tabularnewline
\toprule\noalign{}
\begin{minipage}[b]{\linewidth}\raggedright
Generator
\end{minipage} & \begin{minipage}[b]{\linewidth}\raggedright
Where it stops, per cent of graded plans
\end{minipage} & \begin{minipage}[b]{\linewidth}\raggedright
Attributed to
\end{minipage} & \begin{minipage}[b]{\linewidth}\raggedright
What decides it
\end{minipage} \\
\midrule\noalign{}
\endfirsthead
\toprule\noalign{}
\begin{minipage}[b]{\linewidth}\raggedright
Generator
\end{minipage} & \begin{minipage}[b]{\linewidth}\raggedright
Where it stops, per cent of graded plans
\end{minipage} & \begin{minipage}[b]{\linewidth}\raggedright
Attributed to
\end{minipage} & \begin{minipage}[b]{\linewidth}\raggedright
What decides it
\end{minipage} \\
\midrule\noalign{}
\endhead
\bottomrule\noalign{}
\endlastfoot
House-GAN++ & Proportion check, 28.3 & The generator & RPLAN in
House-GAN++'s own training rendering is refused by the check alone at
5.8 per cent, against 33.7 per cent of its own plans (Section 4.2.2) \\
House-GAN & Compactness, 20.6, then the proportion check, 19.6 &
Proportions: its representation; compactness: not settled & RPLAN
rendered as its training targets is refused by the proportion check
alone at 34.2 per cent, against 32.5 per cent of its own plans; by
compactness alone at 10.2 per cent against 23.2 per cent of its own,
controlled at the unrefitted threshold only (Section 4.2.2;
Supplementary analyses, Section S3) \\
HouseDiffusion & Ingest, 70.4 & Not settled by a control. The overlaps
are substantive: in the 62.7 per cent of plans where two rooms claim the
same floor, the median claimed twice is 1.351 square metres, and 29.8
per cent are refused with no overlap rule at all. This points to the
generator, although no control settles it & Specification, section 9.2;
it contributes no evidence to the margin \\
GSDiff & Right-angle precondition, 83.3, without a verdict & Not
settled. Its segments run off the axes by more than one raster cell, and
rectifying at one cell recovers 32 of 2,425 stops. RPLAN in the same
pass stops there at 0.0 per cent, but no real corpus is graded as the
faces of a wall graph & Restoring the one-cell rectify threshold
recovers 32 of its 2,425 stops (Section 3.2); with its merged living,
dining and foyer class typed, the proportion check alone refuses 7.7 per
cent (Supplementary analyses, Section S4) \\
\end{longtable}

Restoring the one-cell rectify threshold of Section 3.2 recovers
thirty-two of the 2,425 GSDiff plans stopped at the right-angle
precondition. None of the recovered plans then passes, and the change
adds thirty-three build failures.

\begin{figure}
\centering
\includegraphics[width=1\linewidth,height=\textheight,keepaspectratio,alt={One plan from each source drawn at the check where it stops, all at one scale; a plan taller than it is wide is turned a quarter turn. The House-GAN and House-GAN++ panels show the axis-aligned rectilinear polygons the audit traces from each generator's raster room masks, the HouseDiffusion and GSDiff panels draw their vector output as exported (HouseDiffusion's room polygons, GSDiff's faces of its generated wall graph), and the MSD panels draw the partition the audit builds from the dataset's exact geometry, rotated onto each floor's dominant wall direction (Appendix B, decisions 1 and 2). RPLAN plans are not drawn because its terms restrict redistribution. Selection rule, applied without inspecting the plans: for each generator, the first plan in the released archive's order whose first stop under the public checker is that generator's most frequent stop in Tables 2 and 3; for MSD, the worked example of Section 3.1 (floor 1157, apartment u5) and, as a reference, the first apartment in the released order that passes every check (floor 1, apartment u0). Each plan was graded with floorcheck v0.1.1 and its stop asserted before it was drawn. In each panel the highlighted element decides the stop: the room below its threshold at the proportion or compactness check, the floor two rooms both claim for HouseDiffusion's ingest refusal, and the off-axis wall for GSDiff's right-angle precondition.}]{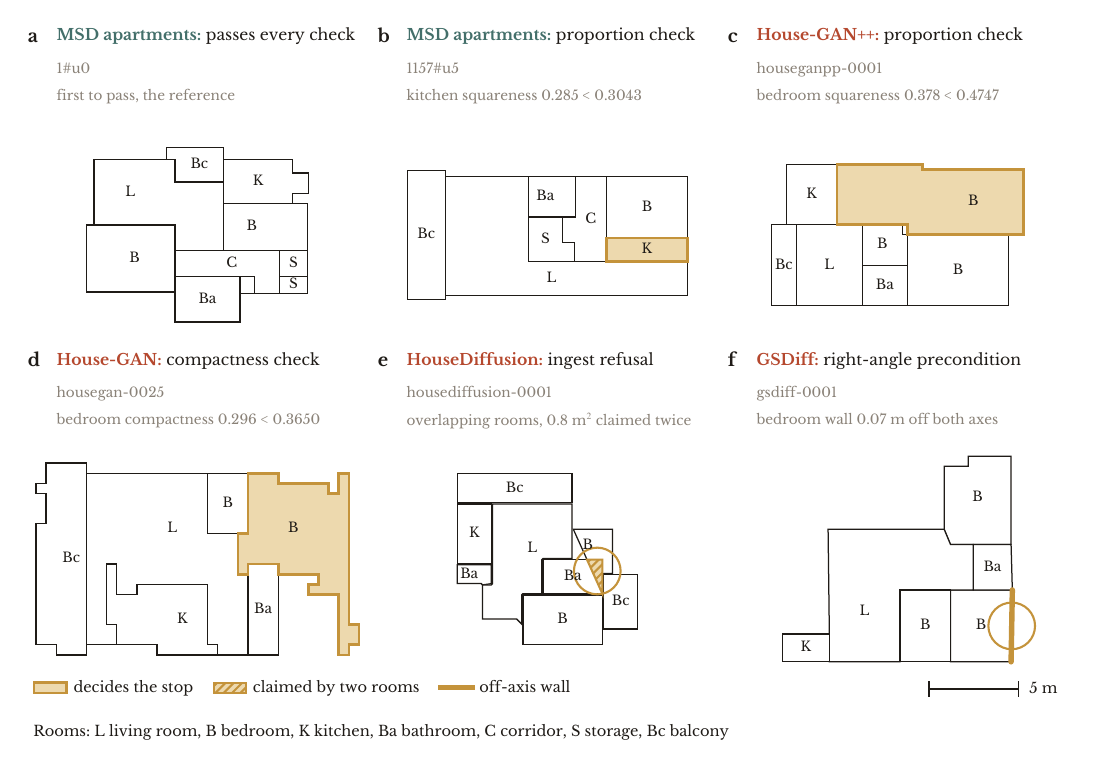}
\caption{One plan from each source drawn at the check where it stops,
all at one scale; a plan taller than it is wide is turned a quarter
turn. The House-GAN and House-GAN++ panels show the axis-aligned
rectilinear polygons the audit traces from each generator's raster room
masks, the HouseDiffusion and GSDiff panels draw their vector output as
exported (HouseDiffusion's room polygons, GSDiff's faces of its
generated wall graph), and the MSD panels draw the partition the audit
builds from the dataset's exact geometry, rotated onto each floor's
dominant wall direction (Appendix B, decisions 1 and 2). RPLAN plans are
not drawn because its terms restrict redistribution. Selection rule,
applied without inspecting the plans: for each generator, the first plan
in the released archive's order whose first stop under the public
checker is that generator's most frequent stop in Tables 2 and 3; for
MSD, the worked example of Section 3.1 (floor 1157, apartment u5) and,
as a reference, the first apartment in the released order that passes
every check (floor 1, apartment u0). Each plan was graded with
floorcheck v0.1.1 and its stop asserted before it was drawn. In each
panel the highlighted element decides the stop: the room below its
threshold at the proportion or compactness check, the floor two rooms
both claim for HouseDiffusion's ingest refusal, and the off-axis wall
for GSDiff's right-angle precondition.}\label{fig:plan_gallery}
\end{figure}

\subsubsection{4.2.2 The proportion check
alone}\label{the-proportion-check-alone}

The proportion check alone refuses 33.7 per cent of the House-GAN++
plans the checker admits, against 5.0 per cent of RPLAN and 4.9 per cent
of the MSD apartments. No measurement control moves that separation by
more than about a point at unchanged thresholds. Refitted on the coarse
corpus, the separation keeps 24.8 to 27.7 of its 28.8 points. This
section shows that the refusal belongs to the generator rather than to
the measurement. Table 4 gives the share of admitted plans that each
shape check refuses when read alone. It is measured with the public
checker's own ingest and geometry at the reported run's thresholds. We
call this the checker's path, as against the solver's path of Table 2.

\LTkeeptrue

\begin{longtable}[]{@{}
  >{\raggedright\arraybackslash}p{(\linewidth - 6\tabcolsep) * \real{0.6118}}
  >{\raggedright\arraybackslash}p{(\linewidth - 6\tabcolsep) * \real{0.1176}}
  >{\raggedright\arraybackslash}p{(\linewidth - 6\tabcolsep) * \real{0.1294}}
  >{\raggedright\arraybackslash}p{(\linewidth - 6\tabcolsep) * \real{0.1412}}@{}}
\caption{The share of admitted plans each shape check refuses when read
alone, on the public checker's path at the reported run's thresholds.
Admitted counts are the checker's; the solver's ingest, on which the
specification counts, admits 14,203 MSD apartments.}\tabularnewline
\toprule\noalign{}
\begin{minipage}[b]{\linewidth}\raggedright
Source
\end{minipage} & \begin{minipage}[b]{\linewidth}\raggedright
Plans the checker's ingest admits
\end{minipage} & \begin{minipage}[b]{\linewidth}\raggedright
Proportion check alone refuses (\%)
\end{minipage} & \begin{minipage}[b]{\linewidth}\raggedright
Compactness check alone refuses (\%)
\end{minipage} \\
\midrule\noalign{}
\endfirsthead
\toprule\noalign{}
\begin{minipage}[b]{\linewidth}\raggedright
Source
\end{minipage} & \begin{minipage}[b]{\linewidth}\raggedright
Plans the checker's ingest admits
\end{minipage} & \begin{minipage}[b]{\linewidth}\raggedright
Proportion check alone refuses (\%)
\end{minipage} & \begin{minipage}[b]{\linewidth}\raggedright
Compactness check alone refuses (\%)
\end{minipage} \\
\midrule\noalign{}
\endhead
\bottomrule\noalign{}
\endlastfoot
House-GAN++ & 3,557 & 33.7 & 2.8 \\
House-GAN & 3,556 & 32.5 & 23.2 \\
HouseDiffusion & 1,189 & 19.8 & 82.2 \\
GSDiff & 2,820 & 6.9 & 3.9 \\
RPLAN ground truth & 3,998 & 5.0 & 5.0 \\
MSD ground truth, per apartment & 14,206 & 4.9 & 4.8 \\
\end{longtable}

\LTkeepfalse

The separation is not a room-count effect. The check alone refuses 5.8
and 4.4 per cent of admitted RPLAN plans of five to six and of seven to
eight rooms. The rates for House-GAN++'s plans are 34.9 and 32.6 per
cent. Nor is the separation an artefact of the label mapping
(Supplementary analyses, Section S4). The generators have no primary
bedroom, so House-GAN++'s counterpart of an RPLAN primary bedroom meets
the stricter Bedroom threshold. This asymmetry accounts for at most 1.4
of its 33.7 per cent, however. Before rounding, the check alone
separates House-GAN++ by 28.8 points from RPLAN and by 28.9 points from
the MSD apartments.

Table 5 tests three measurement objections to that separation.
House-GAN++`s rooms are traced from cells of 0.28125 m, against RPLAN's
0.0703125 m, so the difference might be one of resolution. A spur left
by tracing can stretch a bounding box, and for this reason we apply an
opening to every room of every corpus. Orientation matters as well,
because the unrotated box makes a tilted room look squarer. Of the MSD
apartments' rooms, 3.9 per cent lie more than 2 degrees off the axes,
against 0.4 per cent of House-GAN++'s.

\LTkeeptrue

\begin{longtable}[]{@{}
  >{\raggedright\arraybackslash}p{(\linewidth - 6\tabcolsep) * \real{0.6000}}
  >{\raggedright\arraybackslash}p{(\linewidth - 6\tabcolsep) * \real{0.1294}}
  >{\raggedright\arraybackslash}p{(\linewidth - 6\tabcolsep) * \real{0.1294}}
  >{\raggedright\arraybackslash}p{(\linewidth - 6\tabcolsep) * \real{0.1412}}@{}}
\caption{The separation the proportion check alone draws between
House-GAN++ and each built corpus, in points, under each measurement
control, on the checker's path; ranges in parentheses are over grid
phases, and an empty cell was not measured.}\tabularnewline
\toprule\noalign{}
\begin{minipage}[b]{\linewidth}\raggedright
Control
\end{minipage} & \begin{minipage}[b]{\linewidth}\raggedright
From RPLAN, thresholds unchanged
\end{minipage} & \begin{minipage}[b]{\linewidth}\raggedright
From RPLAN, thresholds refitted
\end{minipage} & \begin{minipage}[b]{\linewidth}\raggedright
From the MSD apartments, thresholds unchanged
\end{minipage} \\
\midrule\noalign{}
\endfirsthead
\toprule\noalign{}
\begin{minipage}[b]{\linewidth}\raggedright
Control
\end{minipage} & \begin{minipage}[b]{\linewidth}\raggedright
From RPLAN, thresholds unchanged
\end{minipage} & \begin{minipage}[b]{\linewidth}\raggedright
From RPLAN, thresholds refitted
\end{minipage} & \begin{minipage}[b]{\linewidth}\raggedright
From the MSD apartments, thresholds unchanged
\end{minipage} \\
\midrule\noalign{}
\endhead
\bottomrule\noalign{}
\endlastfoot
None, geometry as measured & 28.8 & & 28.9 \\
Centre sampling at 0.28125 m, 16 grid phases & 28.0 (27.7 to 28.4) &
26.5 (25.9 to 27.0) & \\
Majority coverage at 0.28125 m, 16 grid phases for RPLAN, 4 for MSD &
28.5 (28.2 to 28.7) & 27.7 (26.2 to 30.6) & 28.7 \\
House-GAN++'s own training rendering at its frame & 28.0 & 24.8 & \\
Opening that deletes every part up to two pixels wide & 28.4 & & 28.1 \\
Minimum-area rectangle in place of the unrotated box & & & 28.7 \\
\end{longtable}

\LTkeepfalse

With the thresholds unchanged, each control moves the separation by
about a point or less. The opening moves it by at most 0.8 points and
the minimum-area rectangle by 0.2 points. Refitted on the coarse corpus,
the separation keeps 24.8 to 27.7 of its 28.8 points. It loses most, 4.0
points, under House-GAN++'s own training rendering. That control works
on clean geometry and does not model noise in the generator's masks
(Supplementary analyses, Section S3). The same rendering is also what a
generator reproducing its training data exactly would emit. RPLAN so
rendered is refused at 5.8 per cent, so House-GAN++'s 33.7 is the
generator's.

The same argument fails for House-GAN. Its grid controls keep most of
its 27.5 points (Supplementary analyses, Table S1). RPLAN rendered as
House-GAN's training targets, bounding boxes with overlaps disjoined, is
refused as often as House-GAN's own plans, however (Table 3).
House-GAN's refusal is therefore not attributable to it beyond its
representation. Neither diffusion generator supports a comparable claim.
HouseDiffusion's compactness figure rests on polygons read as emitted
and supports no claim beyond where it stops. GSDiff's merged living,
dining and foyer class takes no proportion threshold. Held to
LivingRoom's threshold, the strictest of the three, GSDiff is refused at
7.7 rather than 6.9 per cent, close to the built corpora (Supplementary
analyses, Section S4).

What gives the check its discriminating power is the fit per room type.
A single threshold for every room type separates House-GAN++ from RPLAN
by at most 9.0 points at five anchors that need no corpus. At three
anchors it separates them the wrong way, and it refuses far more real
plans. Held to the fitted rate, at a worst-room squareness of 0.209, it
separates them by only 1.6 points (Supplementary analyses, Table S2).
The robust result of this section is therefore the proportion check set
per room type. The controlled shape result concerns House-GAN++, whose
refusal survives every control. House-GAN's refusal is shared with the
representation of its training targets.

A distributional metric on the same feature does not report what the
check reports. Per room type, the Wasserstein distance between room
squareness distributions orders the generators against RPLAN nearly as
the proportion check alone does. Pooled over rooms, however, it
separates them only weakly. This distance is taken within each room type
and averaged over the types, weighted by the generator's share of rooms,
so it is not pooled across room types. It puts House-GAN++ 0.069 from
RPLAN and 0.056 from the MSD apartments, against 0.040 between the two
built corpora, so House-GAN++ lies about 1.4 times as far from the MSD
apartments as RPLAN does. The check, by contrast, refuses House-GAN++ at
about seven times the rate of the MSD apartments, 33.7 against 4.9 per
cent (Table 4). That ratio holds at the declared rate of false rejection
by construction, because the thresholds are fitted to refuse a fixed
share of built plans. Between 87 and 98 per cent of each generator's
distance lies among rooms that clear their thresholds. Four refused
plans in five carry exactly one failing room. Computed per plan on the
worst-room margin, the reduction the check itself uses, the distance
does separate. For House-GAN++ against the MSD apartments, with every
source on RPLAN's thresholds, it is 0.084, against 0.017 between the two
built corpora. With each corpus on the thresholds it is graded on, the
built corpora lie 0.014 apart. Over both built corpora and both sets of
thresholds, House-GAN++ lies about five to seven times as far from a
built corpus as the built corpora lie from each other. This per-plan
distance uses the per-type thresholds through the worst-room margin. The
comparison therefore does not separate the effect of reducing rooms to
plans from the effect of the thresholds.

\subsubsection{4.2.3 The whole-ladder margin and what it rests
on}\label{the-whole-ladder-margin-and-what-it-rests-on}

Over the whole ladder, the margin is the pass rate of the real corpus
that passes least often less that of the generator that passes most
often. These are the MSD apartments and House-GAN++. At the 5 per cent
rate the margin is 12.1, 5.5 and 23.5 points over G, A and V. Table 6
shows what the margin is made of and what it depends on. Each margin is
a net. Over graded plans, the proportion check and the ingest count in
favour of the built corpus. The right-angle precondition and compactness
count against it, and the precondition term is a property of the
representation (Section 4.2.1). The three denominators differ in which
of these stops they keep (Supplementary analyses, Section S3).

\begin{longtable}[]{@{}
  >{\raggedright\arraybackslash}p{(\linewidth - 6\tabcolsep) * \real{0.6660}}
  >{\raggedright\arraybackslash}p{(\linewidth - 6\tabcolsep) * \real{0.1105}}
  >{\raggedright\arraybackslash}p{(\linewidth - 6\tabcolsep) * \real{0.1176}}
  >{\raggedright\arraybackslash}p{(\linewidth - 6\tabcolsep) * \real{0.1059}}@{}}
\caption{The whole-ladder margin, the MSD apartments' pass rate less
House-GAN++'s, in points: its decomposition by check over graded plans,
from Table 2 (the four terms sum to 12.3 from the rounded cells), and
its dependence on the proportion check, the fit and the rate X, the per
cent of real plans each shape check alone refuses. Held-out rows: mean
over 20 splits, RPLAN split by plan and MSD by building, lowest split in
parentheses.}\tabularnewline
\toprule\noalign{}
\begin{minipage}[b]{\linewidth}\raggedright
Condition
\end{minipage} & \begin{minipage}[b]{\linewidth}\raggedright
Graded, G
\end{minipage} & \begin{minipage}[b]{\linewidth}\raggedright
Admitted, A
\end{minipage} & \begin{minipage}[b]{\linewidth}\raggedright
Reached a verdict, V
\end{minipage} \\
\midrule\noalign{}
\endfirsthead
\toprule\noalign{}
\begin{minipage}[b]{\linewidth}\raggedright
Condition
\end{minipage} & \begin{minipage}[b]{\linewidth}\raggedright
Graded, G
\end{minipage} & \begin{minipage}[b]{\linewidth}\raggedright
Admitted, A
\end{minipage} & \begin{minipage}[b]{\linewidth}\raggedright
Reached a verdict, V
\end{minipage} \\
\midrule\noalign{}
\endhead
\bottomrule\noalign{}
\endlastfoot
Reported run, each corpus on its own fit & +12.1 & +5.5 & +23.5 \\
Same, 95 per cent interval with whole MSD buildings resampled & 9.7 to
14.6 & 3.0 to 7.9 & 21.5 to 25.6 \\
Term: proportion check & +24.4 & & \\
Term: ingest refusal & +10.3 & & \\
Term: right-angle precondition & -20.1 & & \\
Term: compactness & -2.3 & & \\
Proportion check evaluated, its rejection suppressed & -13.2 & -23.3 &
-3.4 \\
Compactness check suppressed & +12.8 & & \\
Own fits, seven shared room types & +12.5 & +5.9 & +24.2 \\
Every source on RPLAN's fit & +9.9 & +3.2 & +19.6 \\
Every source on MSD's fit & +10.8 & +4.0 & +22.0 \\
Other corpus's fit & +8.5 & +1.7 & +18.1 \\
Other corpus's compactness threshold only & +9.7 & +3.0 & +19.7 \\
Other corpus's proportion thresholds only & +10.7 & +4.0 & +21.5 \\
Own fits, fitted on one half and graded on the other, X = 5 & +11.4
(+6.6) & +4.7 (-0.3) & +22.8 (+17.2) \\
Same, X = 1 & -2.9 (-6.1) & -12.0 (-15.2) & +9.6 (+7.0) \\
Same, X = 2 & +3.5 (-0.6) & -4.7 (-8.9) & +16.1 (+11.3) \\
Same, X = 10 & +17.1 (+12.6) & +11.8 (+7.2) & +26.0 (+20.6) \\
Same, X = 15 & +18.5 (+14.9) & +14.0 (+10.5) & +25.4 (+21.0) \\
Same, X = 20 & +18.5 (+15.5) & +14.7 (+11.7) & +23.6 (+19.4) \\
Same, X = 30 & +16.7 (+12.0) & +14.0 (+9.2) & +19.4 (+14.2) \\
\end{longtable}

Leaving the precondition stops out of V does not flatter the built
corpus. The checks run in order, so an apartment the precondition stops
has already passed both shape checks. Of the 3,225 admitted apartments
whose wall loops fail the precondition, either shape check alone refuses
10.9 per cent, against 8.2 of the others. Those apartments are rejected
first and stay in V as failures. V therefore drops only the passing
apartments of that group. Without the precondition, the MSD apartments
would pass 91.2 per cent of admitted plans on the checker's path. This
gives a margin of +25.8 against the checker's +23.6 over V, so V
understates the margin rather than overstating it.

Room count also runs against the MSD apartments (Section 3.2), since a
shape check judges a plan by its worst room. Apartments on one MSD floor
also share a designer and its conversion artefacts. The margin's
intervals therefore resample whole buildings, inferred from shared
structure (Supplementary analyses, Section S3). These intervals are
about 1.2 to 1.4 times as wide as intervals that treat plans as
independent. Other intervals treat plans as independent unless stated.

Which source passes more often depends on the proportion check. With
that check evaluated but its rejection suppressed, the margin inverts on
every denominator. Over graded plans it falls to -13.2 (interval -14.4
to -11.9), and House-GAN++ passes 86.5 per cent of its graded plans
against the MSD apartments' 73.3. The released outcome command
reproduces this inversion on the public checker's verdicts, at -13.1,
-23.2 and -3.4. Switching off the compactness threshold instead barely
moves the margin. House-GAN then rises from 48.7 to 60.1 per cent,
however, and overtakes House-GAN++. Its compactness refusal is
controlled only at the unrefitted threshold.

A fitted threshold invites two objections. The first is that the check
might pass a corpus because that corpus wrote it. The exchanged, common
and shared-type fits of Table 6 test this objection. In these fits, a
room whose type the other corpus does not label takes no proportion
threshold. The second is that the graded plans might leak into their own
thresholds, which the held-out fits test. In every exchanged row the 95
per cent interval excludes zero, with one exception. Over admitted plans
on the other corpus's fit, its lower end is at zero, and the interval
spans zero, -0.8 to +4.1, when whole buildings are resampled. The
counterfactual keeps its sign on every denominator under building
resampling. No exchange inverts the margin. The pair that sets the
margin is also the same in every row, held-out included, on every
denominator. It is the MSD apartments, the built corpus that passes
least, against House-GAN++, the generator that passes most (Figure 6).
The full exchange narrows the graded margin by about a third, mostly
through compactness, whose MSD threshold is the lower. It also lifts
RPLAN to 95.3 per cent of graded plans, which still does not bind. The
margin is therefore not a tautology, though the transfer is lossy and
corpus-specific. The proportion check is an empirical shape threshold,
not a validity check, which nothing public derives. Nor is it a pure
corpus-conformance measure, since its margin survives the other corpus's
proportion thresholds with an interval that excludes zero on every
denominator.

\begin{figure}
\centering
\includegraphics[width=0.75\linewidth,height=\textheight,keepaspectratio,alt={The margin over graded plans under each condition of Table 6, with 95 per cent intervals; the held-out fit shows its mean and range over 20 splits. Top: the reported run and the two counterfactuals. Middle: exchanged, shared-type and held-out fits. Bottom: the checker's own verdicts. Left of the dashed line, a generator passes more often than a built corpus.}]{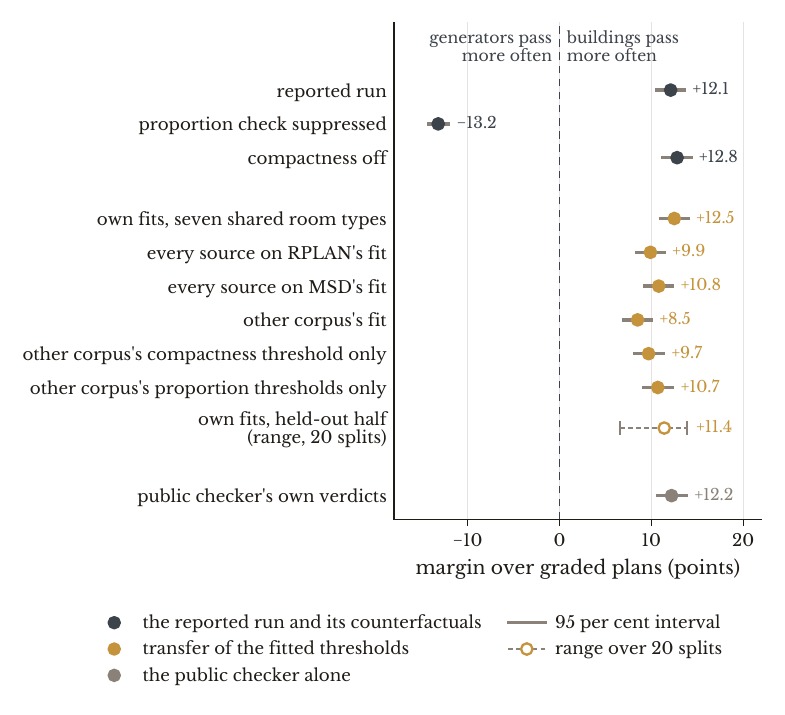}
\caption{The margin over graded plans under each condition of Table 6,
with 95 per cent intervals; the held-out fit shows its mean and range
over 20 splits. Top: the reported run and the two counterfactuals.
Middle: exchanged, shared-type and held-out fits. Bottom: the checker's
own verdicts. Left of the dashed line, a generator passes more often
than a built corpus.}\label{fig:margin_counterfactuals}
\end{figure}

The lowest swept rate at which the margin is positive in every held-out
split is X = 5 over G, X = 10 over A and X = 1 over V (Appendix C).
Snapping MSD's room geometry to a grid would raise the margin over G and
A by about nineteen points through hidden precondition stops, so we use
the exact geometry (Appendix C).

The proportion threshold does not come from a building code. If the
code's own shape constraint, IRC R304.2's least horizontal dimension of
7 feet, is put in its place, the margin is smaller. At the scale and
width threshold we infer for the generators' frames, it reverses over
admitted plans. One frame step of the threshold removes the reversal,
however (Supplementary analyses, Section S3).

\subsection{4.3 What ground truth fails}\label{what-ground-truth-fails}

The corpora the field trains and evaluates against do not pass the
ladder either. Each shape check alone refuses 5 per cent of real plans
by construction. Refitted at X = 5 and graded on held-out halves, the
whole ladder does not pass 8.8 per cent of RPLAN plans and 30.2 per cent
of MSD apartments, the mean over 20 splits. This population is distinct
from the 3,998 held-out RPLAN plans of Section 4.1. The MSD share is not
about room shape. Over the full corpus, the right-angle precondition
stops 20.1 of the 29.7 per cent of graded MSD apartments that do not
pass (Table 2). It fails on 3,225 of the 14,206 MSD apartments in A,
22.7 per cent, whatever the thresholds. For these apartments the audit
declines to speak rather than refusing a building. The 3,225 are every
admitted apartment with a non-orthogonal wall loop on the checker's
path. Table 2's 2,879, on the solver's path, count only those the
precondition stops. On the checker's path, 350 of the 3,225 failed a
shape check first. The room-level rule set aside in Section 3.1 would
refuse far more. Under it, the two shape checks together would refuse
45.3 per cent of RPLAN plans. Figure 7 draws MSD rooms either side of
each threshold.

\begin{figure}
\centering
\includegraphics[width=1\linewidth,height=\textheight,keepaspectratio,alt={MSD rooms drawn to one scale either side of the two corpus-fitted thresholds of the reported run. (a) Compactness, with thresholds 0.365 (RPLAN apartments) and 0.3237 (MSD apartments). (b) Proportions, against each room type's MSD threshold. Solid outlines pass, dashed outlines fall below. Rooms are not a random sample; RPLAN's licence forbids drawing its rooms.}]{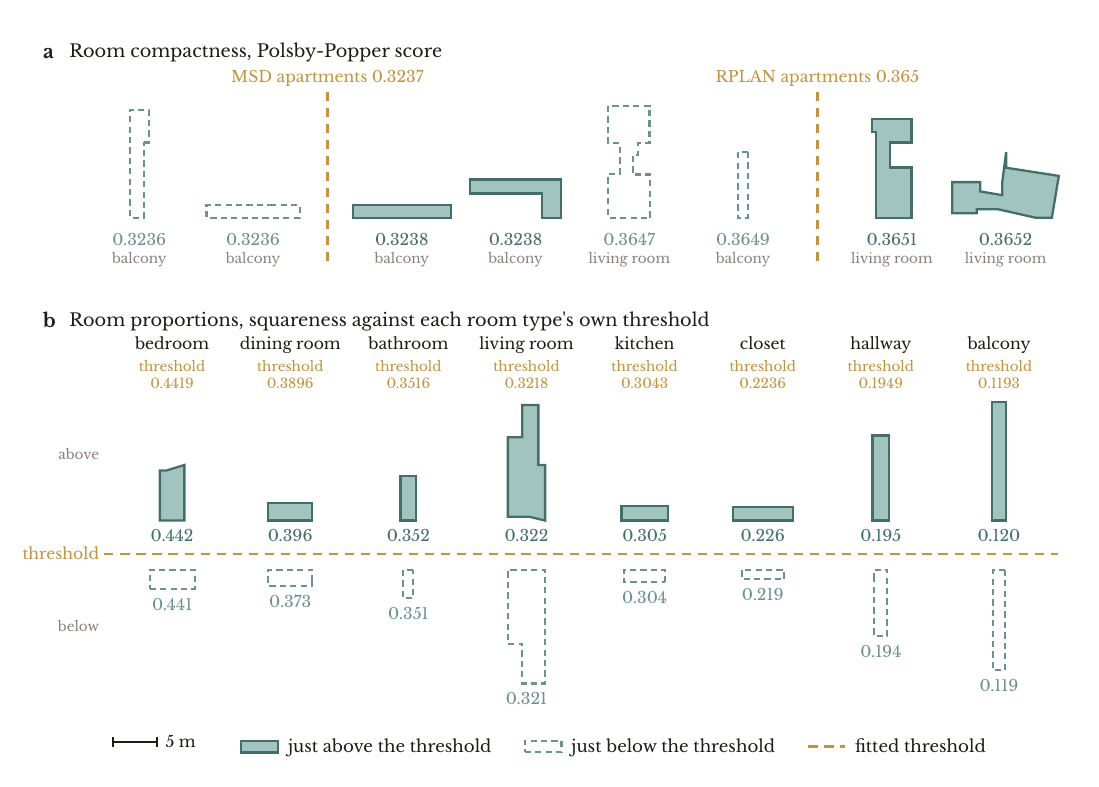}
\caption{MSD rooms drawn to one scale either side of the two
corpus-fitted thresholds of the reported run. (a) Compactness, with
thresholds 0.365 (RPLAN apartments) and 0.3237 (MSD apartments). (b)
Proportions, against each room type's MSD threshold. Solid outlines
pass, dashed outlines fall below. Rooms are not a random sample; RPLAN's
licence forbids drawing its rooms.}\label{fig:rooms_at_the_floors}
\end{figure}

Four readings of these refusals apply, and none is tested here. The
ladder may be stricter than practice. The corpora may carry conversion
artefacts, although snapping rooms to a grid made the checks pass rather
than fail (Section 4.2.3). Real plans may contain rooms an architect
would flag but build. This is the likeliest reading for the proportion
check. Finally, a threshold fitted to a rate refuses that rate of its
own corpus by construction, so the ground-truth rate is not a finding
about buildings.

\section{5. Discussion}\label{discussion}

\subsection{5.1 What a generator paper should
report}\label{what-a-generator-paper-should-report}

Compatibility does not say where a generator's plans stop. We recover it
from the audited plans. Supplementary analyses, Section S3, gives the
procedure and explains why the values are not comparable with published
ones. The two generators that lead on Compatibility, GSDiff (1.25) and
HouseDiffusion (2.68), pass least (Table 2). House-GAN++ (4.19) and
House-GAN (4.25) show the opposite.

House-GAN and House-GAN++ produce almost all generated plans the ladder
passes. Within them, the area under Compatibility's ROC curve for
predicting a pass lies between 0.43 and 0.53 over G and over A, both
pooled and within one requested room count. The 95 per cent bootstrap
intervals lie between 0.41 and 0.55 (Supplementary analyses, Table S4).
A graph measure has no reason to track a check of room shape. This
matters, however, because Compatibility is reported as a quality score.
Within these generators it carries no consistent signal about where
their plans stop. The intervals exclude 0.5 in six of the eight cells,
but on both sides. A weak to moderate relation appears only within the
two diffusion generators, whose few passing plans match their requests
better than their failing plans do. We therefore recommend four
quantities beside Compatibility, each computed by the released tools
without the private solver.

\emph{The per-check refusal profile.} A single pass rate hides where
plans fail, and each place the audited generators fail points to a
different repair (Section 4.2.1). A precondition stop belongs in the
profile as a place where plans stop, not as a rejection.

\emph{The margin over ground truth on the same checks.} The real corpora
fail the ladder too. A pass rate therefore means something only beside a
built corpus graded the same way, with an interval that resamples whole
buildings (Section 4.2.3).

\emph{The false-rejection rate of the thresholds.} The margin moves with
the rate (Table 6), so pass rates at unstated rates are not comparable.
A threshold read from the room distribution states no rate at all.

\emph{The share with no verdict.} Plans refused at ingest, stopped at a
precondition or failing to build are not rejections, and the denominator
moves the margin (Section 4.2.3).

\subsection{5.2 Using calibrated checks beyond this
audit}\label{using-calibrated-checks-beyond-this-audit}

The audit shows that ordered checks returning a named rejection reason
can evaluate generative models. It also shows that they report what a
distance to the request does not. They report where each generator's
plans stop, and so which part of a model or its output representation a
repair should target. The protocol is not tied to this solver. Where no
external standard fixes a threshold, real samples can calibrate it at a
declared rate. Each threshold then states something empirical about a
corpus rather than a norm. The rate must then be reported beside the
margin it moves (Section 4.2.3). The real samples the checks refuse
(Section 4.3) are the visible cost. A transfer specification records, as
this one does (Section 3.2 and Appendix B), which checks a corpus can
carry and the source of each threshold. This paper followed the route
for one solver, with public corpora in place of a practice's own plans.

The same checks could also enter training, as filters on training data
or on samples. They could also serve as verifiable rewards of the kind
\citet{lara2026rlvr} optimise, whose rewards they concede are necessary
but not sufficient. We have not trained with them, and two cautions
follow from what was measured. A generator optimised against the ladder
would be graded by the thresholds it was trained on. Its pass rate would
then no longer be read against an independent check. A threshold fitted
at a stated rate also refuses that share of built plans by construction
(Section 4.3). A filter at that threshold would therefore discard real
plans as well as generated ones.

\subsection{5.3 Limitations}\label{limitations}

The ladder is residential. Its thresholds are fitted on apartments and
single floors of apartment buildings, and nothing here grades another
building type.

The two shape thresholds are fitted to the corpora they grade, so every
margin in Section 4.2 depends on the fit. Section 4.2.3 measures how far
the margin moves under other fits but does not bound it. Held out, the
margin is lower on all three denominators by 0.7 to 0.8 points.
Exchanged, the graded margin loses up to a third. These movements are
not a correction, and the intervals do not cover them. Even the
clustered intervals may be somewhat too narrow, since the inferred
building key can split one building into several.

Grading MSD on its exact geometry leaves a fifth of its apartments
without a verdict. The grid conversion that would avoid it inflates the
margin (Section 4.2.3).

Each disagreement between the checker and the solver is assigned to one
mechanism. The largest class is attributed chiefly to a quantisation
step on which the two implementations differ by design (Supplementary
analyses, Section S2). The measurement controls of Section 4.2.2 test
the check alone on the checker's geometry, not the solver's ladder.

The proportion threshold does not come from a building code, and the
code's own shape constraint in its place gives a smaller margin (Section
4.2.3).

The checks an architect reaches for first are not in this ladder
(Section 3.1). They are corridor topology, a breach of the boundary
between dwellings and door clear width. No expert validated the two
shape thresholds. The paper therefore claims that the proportion
threshold carries the margin between the built corpora and the generator
that passes most often. It does not claim that the threshold is a
validity criterion. Calling the ladder a test of validity would claim
more than was measured. Passing is not shown to mean acceptance in
practice, and the shape thresholds are empirical, drawn from a corpus
rather than from practice. The ladder as a whole is a proxy for
professional judgment, not a model of it. Passing every check does not
establish that a plan is good architecture.

\section{6. Conclusion}\label{conclusion}

Compatibility became the field's main measure because it was cheap,
automatic and defined without reference to any building. On RPLAN it is
approaching saturation. On RQ1, the checks of a production layout solver
can be carried to public floor plan data in part: three of the six run.
They transfer in their form, with shape thresholds fitted to refuse 5
per cent of built plans. A public checker reproduces the solver's first
failing check on 100.0 per cent of 3,998 held-out RPLAN plans (Section
4.1). On RQ2, the generators stop at different checks. The robust result
is the proportion check, set per room type, which alone separates
House-GAN++ from both built corpora (Section 4.2.2). The whole-ladder
margin rests on that check. At the 5 per cent rate it holds in every
held-out split over G and V. Over A it holds on the mean of the splits
but not in every split, and it holds in every split from 10 per cent
(Section 4.2.3). On RQ3, within the two generators that pass most often,
Compatibility as recovered from the plans separates passing from failing
plans scarcely at all. Its area under the ROC curve lies within 0.07 of
chance in either direction (Section 5.1).

The paper establishes two things. First, ordered checks with named
rejection reasons can grade the generated plans they can read and name
where each stops, which the standard metric cannot. The checks are
calibrated on real samples at a declared rate and applied to those
samples as well. Second, a generator paper can report, beside
Compatibility, where its plans stop, its margin over ground truth, the
rate its thresholds are set at and its share with no verdict. No step of
the protocol is specific to floor plans. Other domains also generate
samples that must meet constraints no standard quantifies, among them
the mobile interfaces of Rico \citep{deka2017rico} and the document
pages of PubLayNet \citep{zhong2019publaynet}. There, thresholds can be
calibrated on real samples at a declared rate, and the real samples
graded by the same checks. Whether a fitted threshold transfers between
corpora can be checked in the same way.

\section{Declarations}\label{declarations}

\subsection{Declaration of competing
interest}\label{declaration-of-competing-interest}

The author is employed by Skidmore, Owings \& Merrill (SOM), which
develops the production layout solver whose acceptance checks this study
transcribes and which served as the reference implementation. The author
declares no other competing financial interests or personal
relationships that could have appeared to influence the work reported in
this paper.

\subsection{CRediT authorship contribution
statement}\label{credit-authorship-contribution-statement}

Yiliang Shao: Conceptualization, Methodology, Software, Formal analysis,
Investigation, Data curation, Visualization, Writing - original draft,
Writing - review and editing.

\subsection{Funding}\label{funding}

This research did not receive any specific grant from funding agencies
in the public, commercial, or not-for-profit sectors.

\subsection{Data availability}\label{data-availability}

The checker, the export and outcome tools, the converted MSD documents
and the generated plans' geometry are released as floorcheck v0.1.1,
https://github.com/yiliangs/floorcheck, archived at
https://doi.org/10.5281/zenodo.23129970 \citep{shao2026floorcheck}. The
checker code is under the MIT licence, the converted MSD documents under
CC BY 4.0 as MSD requires, and the generated plans' geometry under CC
BY-NC 4.0, chosen to respect the generators' non-commercial terms.
RPLAN's licence forbids redistributing documents derived from it. The
generated plans are therefore deposited in reduced form, as room types
and polygons only, and their RPLAN plan identifiers and requested graphs
are withheld. Under the public checker each deposited plan still reaches
the verdict its full document reached, and the generators' rows of the
checker's outcome table are reproduced exactly. MSD is available from
4TU.ResearchData under CC BY 4.0, and RPLAN from its authors on request.
The production solver is proprietary and is not available.

\subsection{Supplementary material}\label{supplementary-material}

Two supplementary files accompany this paper. Supplementary material 1,
Transfer specification: the solver's checks as transferred to public
floor plan data, with the transfer decision for each check and the
source of each threshold. Supplementary material 2, Supplementary
analyses: reproduction instructions for MSD, the triage of the oracle
disagreements, supporting measurements and the room labels and threshold
types, cited in the text by section (S1 to S4) and table (Tables S1 to
S4).

\subsection{Declaration of generative AI and AI-assisted technologies in
the manuscript preparation
process}\label{declaration-of-generative-ai-and-ai-assisted-technologies-in-the-manuscript-preparation-process}

During the preparation of this work the author used Claude (Anthropic)
to assist with drafting and editing the text. After using this tool, the
author reviewed, tested and edited the content as needed and takes full
responsibility for the content of the publication.

\bibliography{references}

\section{Appendix A. Where each constant comes
from}\label{appendix-a.-where-each-constant-comes-from}

Table 7 lists the constants that the transferable checks and the ingest
read on a public plan. Lengths are in metres, one RPLAN cell being
0.0703125 m, and where a decision of Appendix B applies it overrides the
tabled derivation. No constant in force comes from the production solver
except its cap on a bedroom's proportion threshold, 0.60, which binds on
no fitted threshold (Section 3.1); the checks that do not transfer, and
the requirements the solver applies when it builds a plan, are described
in Sections 8.5 to 8.7 of the accompanying specification. Two of the
constants mean the reverse of what the solver's interface names invite:
the proportion constant is a minimum of short side over long, so a
larger value is stricter, and the area requirement is a proportional
weight, not a minimum.

\begin{longtable}[]{@{}
  >{\raggedright\arraybackslash}p{(\linewidth - 6\tabcolsep) * \real{0.2118}}
  >{\raggedright\arraybackslash}p{(\linewidth - 6\tabcolsep) * \real{0.2745}}
  >{\raggedright\arraybackslash}p{(\linewidth - 6\tabcolsep) * \real{0.3020}}
  >{\raggedright\arraybackslash}p{(\linewidth - 6\tabcolsep) * \real{0.2116}}@{}}
\caption{The constants the transferable checks and the ingest read on a
public plan, with the anchor of each; lengths in metres.}\tabularnewline
\toprule\noalign{}
\begin{minipage}[b]{\linewidth}\raggedright
Constant
\end{minipage} & \begin{minipage}[b]{\linewidth}\raggedright
What it governs
\end{minipage} & \begin{minipage}[b]{\linewidth}\raggedright
Anchor
\end{minipage} & \begin{minipage}[b]{\linewidth}\raggedright
Value
\end{minipage} \\
\midrule\noalign{}
\endfirsthead
\toprule\noalign{}
\begin{minipage}[b]{\linewidth}\raggedright
Constant
\end{minipage} & \begin{minipage}[b]{\linewidth}\raggedright
What it governs
\end{minipage} & \begin{minipage}[b]{\linewidth}\raggedright
Anchor
\end{minipage} & \begin{minipage}[b]{\linewidth}\raggedright
Value
\end{minipage} \\
\midrule\noalign{}
\endhead
\bottomrule\noalign{}
\endlastfoot
\texttt{joinTol} & Endpoint distance at which wall segments chain & Half
an RPLAN cell & 0.03515625 \\
\texttt{rectifyThreshold} & Deviation from horizontal or vertical that
the construction step snaps straight & One RPLAN cell; superseded by
decision 5 of Appendix B & 0.0703125 derived; 0 in force \\
\texttt{abstol} & Short-segment and duplicate-vertex removal; the
identity quantisation & One millimetre, about one seventieth of an RPLAN
cell & 0.001 \\
\texttt{angtol}, rad & Collinearity angle when simplifying a polygon &
One cell of staircase over one door width, atan(0.0703125 / 0.813) &
0.0862706 \\
\texttt{localtol} & Whether a span lies on a boundary & One RPLAN cell &
0.0703125 \\
\texttt{minPassingPPScore} & Polsby-Popper compactness threshold &
Corpus-fitted, no external validation: no statutory or peer-reviewed
threshold; a square scores 0.7854, a 2:1 rectangle 0.6981, a 3:1
rectangle 0.5890 & Plan-level fit at a 5 per cent false rejection: 0.365
RPLAN apartments, 0.3237 MSD apartments, 0.2406 MSD floors (reported
run) \\
\texttt{propReq}, per room type & A minimum squareness, short side over
long & Corpus-fitted, no external validation: no code sets a room aspect
ratio & One quantile of each type's squareness per kind at the same
rate; for RPLAN apartments bedroom 0.4747, bathroom 0.3413, kitchen
0.2748, living room 0.3486 (reported run) \\
Bedroom cap & The highest proportion threshold enforced for a bedroom,
whatever its \texttt{propReq} & The solver's behaviour, applied by the
released checker as the solver does & 0.60 \\
\texttt{minWidth} & Least horizontal dimension, not on the ladder
(Section 4.2.3) & IRC 2021 R304.2, 7 ft in any horizontal dimension &
2.134 \\
\end{longtable}

The released checker stores the two corpus-fitted thresholds in
\texttt{plan-level-floors.json}, where floor means a lower bound, not a
storey. For each kind of plan, the compactness threshold is the 5 per
cent quantile, over that kind's admitted real plans, of each plan's
least compact non-hallway room. Every room type with a proportion
threshold takes the same quantile p of its own rooms' squareness, and
bisection chooses p so that the proportion check alone refuses 5 per
cent of that kind's plans; the bedroom cap applies after the fit.

\section{Appendix B. Per-check transfer
decisions}\label{appendix-b.-per-check-transfer-decisions}

Table 8 lists, in pipeline order, the choices where derivation alone
could not settle how a check transfers to public data, each with the
section of the accompanying specification that sets it out.

\LTkeeptrue

\begin{longtable}[]{@{}
  >{\raggedright\arraybackslash}p{(\linewidth - 4\tabcolsep) * \real{0.0353}}
  >{\raggedright\arraybackslash}p{(\linewidth - 4\tabcolsep) * \real{0.8000}}
  >{\raggedright\arraybackslash}p{(\linewidth - 4\tabcolsep) * \real{0.1647}}@{}}
\caption{The per-check transfer decisions, in pipeline order, with the
section of the specification that sets each out.}\tabularnewline
\toprule\noalign{}
\begin{minipage}[b]{\linewidth}\raggedright
\#
\end{minipage} & \begin{minipage}[b]{\linewidth}\raggedright
Decision
\end{minipage} & \begin{minipage}[b]{\linewidth}\raggedright
Specification section
\end{minipage} \\
\midrule\noalign{}
\endfirsthead
\toprule\noalign{}
\begin{minipage}[b]{\linewidth}\raggedright
\#
\end{minipage} & \begin{minipage}[b]{\linewidth}\raggedright
Decision
\end{minipage} & \begin{minipage}[b]{\linewidth}\raggedright
Specification section
\end{minipage} \\
\midrule\noalign{}
\endhead
\bottomrule\noalign{}
\endlastfoot
1 & MSD floors are divided into rooms from their exact geometry rather
than from a grid-snapped conversion & 5.6 \\
2 & The exact-geometry conversion welds features up to 0.40 m, the
widest wall the corpus draws & 5.6 \\
3 & A plan whose rooms do not tile the plan's outline is refused at
ingest and counted apart & 6 \\
4 & An open gap between rooms that do not touch is closed by the same
rule as an enclosed void, for every source & 6.3 \\
5 & The rectify threshold is zero, decided when one RPLAN cell refused
65 per cent of House-GAN++ plans inside the straightening step & 8.8 \\
6 & The compactness threshold is fitted per kind so that the check alone
refuses 5 per cent of real plans & 8.2 \\
7 & The proportion thresholds are one quantile per kind fitted to the
same rate & 8.3 \\
8 & Check 6 is not transferable, since it re-tests only a requested
adjacency program, and the checker does not implement it & 8.6 \\
9 & A whole MSD floor graded as one plan is excluded from the margins,
since failure compounds with room count & 8.8 \\
10 & Oracle agreement is scoped to finished plans, not to the solver's
build rate & 9.3 \\
\end{longtable}

\LTkeepfalse

The rectify threshold was set to zero when, at the publicly derived
value of one RPLAN raster cell, the solver failed to build 65 per cent
of House-GAN++ plans. This share was measured once, at an earlier
commit, and not remeasured.

\clearpage

\section{Appendix C. The rate sweep and the grid
conversion}\label{appendix-c.-the-rate-sweep-and-the-grid-conversion}

The rate of false rejection itself is swept in Table 6's held-out rows.
The whole ladder's refusal of held-out real RPLAN plans climbs with it,
from 1.8 per cent at X = 1 to 8.8 at X = 5 and 47.4 at X = 30. At X = 1,
with both shape checks nearly off, House-GAN++ passes more often than
the MSD apartments over G and A, but not over V. That reversal at low
rates comes from the right-angle precondition, which counts against the
built corpus on G and A only, and not from room shape. The 5 per cent
rate is the convention of Section 3.1 rather than a point chosen from
the curve. At that rate, the whole-ladder margin holds in every held-out
split over G and V. Over A it holds on the mean of the splits, +4.7, but
not in every split, and the lowest split falls to -0.3. Over A it holds
in every split from 10 per cent. Its conditions are thus the rate and
the denominator, and it rests on the proportion check (Section 4.2.2).
The proportion check's separation (Section 4.2.2), by contrast, is
measured on admitted plans and nets no precondition term.

Snapping MSD's room geometry to a grid would inflate the margin. Such a
conversion resolves every floor onto a 0.38 m grid, with a median
per-room width error of 0.147 m. It also supplies the percentiles its
own plans are judged against. The MSD apartments were converted on that
grid and graded by the public checker. On thresholds fitted to the exact
geometry, the margin over House-GAN++ is then +31.0, +24.8 and +24.8
points over G, A and V. With MSD's thresholds refitted on the grid, it
is +31.8, +25.6 and +25.6. The checker gives +12.2, +5.6 and +23.6 on
the exact geometry. Every grid edge lies on an axis, so the grid stops
no apartment at the right-angle precondition. The conversion therefore
moves the margin over G and A by about nineteen points, almost all of
them hidden precondition stops. Over V, which those stops do not enter,
it moves the margin by only 1.2 to 2.0. We use the exact geometry. The
cost is that the precondition stops a fifth of the MSD apartments short
of a verdict. Grading per floor also refuses 138 floors whose rooms
close into more than one part.

\section{Supplementary analyses}\label{supplementary-analyses}

Section, table, figure and appendix numbers without the prefix S refer
to the paper.

\setcounter{table}{0}\renewcommand{\thetable}{S\arabic{table}}
\setcounter{figure}{0}\renewcommand{\thefigure}{S\arabic{figure}}

\subsection{S1. Reproduction instructions for
MSD}\label{s1.-reproduction-instructions-for-msd}

Everything up to the fitted thresholds and the plans in the checker's
schema can be rebuilt from the public dataset and graded with the
checker, whose agreement with the solver on MSD was measured on the
plans it was corrected against, only RPLAN having a held-out population
(Section 4.1); the cells as the solver computes them need the private
solver. The commands below run from the root of the released archive,
every module they name being in it. RPLAN's licence forbids
redistribution. MSD is released on 4TU.ResearchData under CC BY 4.0.
From the archive root, download into \texttt{data/msd/}:

\begin{verbatim}
BASE=https://data.4tu.nl/file/e1d89cb5-6872-48fc-be63-aadd687ee6f9
curl -L -o data/msd/modified-swiss-dwellings-v1-train.zip \
  "$BASE/279ef4b4-d3bd-41f4-b0c9-5e9af8cce6f6"
unzip data/msd/modified-swiss-dwellings-v1-train.zip \
  -d data/msd/train_extracted
\end{verbatim}

The MSD loader divides each floor into rooms on every load under the
decisions in Appendix B. Export the plans, fit the plan-level
thresholds, grade with the checker, and tabulate the four quantities of
Section 5.1 over the checker's verdicts; the room-type statistics the
checker grades against ship fitted, as
\texttt{floorcheck/category-statistics.md}:

\begin{verbatim}
P=data/derived/panel
python -m tools.export_plans --msd 4167 --msd-units 4000 --out $P/plans
python -m tools.export_plans --rplan 4000 --rplan-stride 20 \
  --rplan-path /path/to/data.mat --out $P/plans
for g in housegan=4000 houseganpp=4000 housediffusion=3991; do
  python -m tools.export_plans --baseline $g --out $P/plans
done
python -m tools.msd_buildings --root data/msd/train_extracted/struct_in \
  --cache data/derived/msd-buildings.json
python -m tools.plan_level_floor $P/plans --workers 8 --rebuild \
  --buildings data/derived/msd-buildings.json \
  --features data/derived/plan-level-features.jsonl \
  --emit floorcheck/plan-level-floors.json \
  --out plan-level-floor.gen.md --json plan-level-floor.json
for f in $P/plans/msd_units/*.json; do
  floorcheck "$f" --requirements dataset-p05
done
python -m tools.checker_outcomes $P/plans --profile \
  --buildings data/derived/msd-buildings.json --out checker-outcomes.md
\end{verbatim}

\texttt{-\/-msd} and \texttt{-\/-msd-units} set how many MSD floors are
exported, as whole floors and as their apartments;
\texttt{-\/-rplan-stride\ 20} takes every twentieth RPLAN plan from the
reader's own copy of RPLAN's \texttt{data.mat}, named by
\texttt{-\/-rplan-path}; \texttt{-\/-baseline\ NAME=N} exports N samples
of a generator from its directory under \texttt{data/samples/}, which
the reader fills from each generator's own release. \texttt{-\/-rebuild}
reads every plan afresh rather than from a cached feature file,
\texttt{-\/-emit} writes the fitted thresholds the checker grades
against to \texttt{plan-level-floors.json} (the code's names use floor
for a fitted threshold, a lower bound, never for a storey),
\texttt{-\/-requirements\ dataset-p05} grades against them,
\texttt{-\/-profile} adds where each plan stopped and the share with no
verdict, and \texttt{-\/-buildings} adds the margin with whole MSD
buildings resampled.

The remaining results need the private solver and were computed in the
working repository from which the archive is cut: the solver graded the
audited plans for every cell of Table 2, for its figures of the
counterfactuals of Section 4.2.3, which the released outcome command
reproduces on the checker's verdicts within 0.1 point, and for the
exchanged fits of Section 4.2.3, with the room-type statistics refit
there; the intervals, whether plans or whole buildings are resampled,
and the agreement of Section 4.1 were tabulated from the solver's
per-plan listings. Every setting is recorded with its cells, but none of
those cells can be rerun without the solver.

The export held 4,028 floors, the 4,167 graded less the 138 that grading
per floor refuses and floor 2118, which the loader cannot read, and
14,323 MSD apartments, one below the reported run's 14,324, since the
apartment on floor 2118 has no exported document. The solver's cells
come from the audit's grading driver. GSDiff is graded in a separate
pass; its plans are exported by the same exporter with
\texttt{-\/-baseline\ gsdiff=100000}, a cap above the 2,911 of its
split, into a directory of their own and graded with \texttt{floorcheck}
as above, and \texttt{tools.oracle\_agreement} pairs the checker's
verdicts on them with the solver's listing for Section 4.1; the outcome
command keeps GSDiff out of its margin table.

\subsection{S2. Oracle-disagreement
triage}\label{s2.-oracle-disagreement-triage}

On the reported run the two ingests agree on 99.9 per cent of the 34,342
plans for which both sides hold a document, Table 2's 34,482 graded
outside GSDiff less the 139 MSD floors and one apartment that have no
exported document (Section S1). Agreement is scoped to finished plans
(Appendix B, decision 10), so of the 30,472 plans both ingests accept,
the 135 the solver failed to build with no precondition stop on either
side are left out. The remaining 30,337 count as a disagreement every
plan that one side stops and the other does not, 59 of them at the
precondition; the 26,126 that both carry to a verdict leave out every
stop, 4,152 of them at the precondition on both sides.

On these populations the checker and the solver agree on the first
failing check for 30,221 of the 30,337 plans both ingests accept. They
agree on whether a plan passes for 99.8 per cent of the 26,126 both
carry to a verdict (99.7 to 99.9). By source, agreement on the first
failing check runs from 99.5 per cent on HouseDiffusion to 100.0 per
cent on RPLAN. GSDiff, a separate population, agrees on 100.0 per cent
of the 376 plans both sides carry to a verdict (99.0 to 100.0). These
plans are few because the precondition stops most of its plans first.

On the reported run, leaving aside the stops at the right-angle
precondition and the plans the solver failed to build, the checker and
the oracle disagree on 95 plans, each assigned to exactly one mechanism
tested in a fixed order, the first that matches; three classes occur,
and no check 4 stop is compared, since every one is a failure of the
precondition. In 53 the two sides measure the same score on different
rings of the same room, a ring being the closed outline each side's
ingest produces for a room; in 38 the two ingests disagree about the
plan, so there is no check to compare; and in 4 the two quantisation
steps differ, the disagreement disappearing when the checker is handed
the quantised geometry the solver saw. Of the 4,211 plans stopped at the
right-angle precondition on at least one side, 2,887 MSD apartments,
1,205 MSD floors and 119 HouseDiffusion plans, 4,152 are stopped on both
sides; those stopped on one side only, 19 by the solver and 40 by the
checker, are left out of the 26,126 with every other stop, and 33 of
them are disagreements about whether the precondition holds and the
other 26 plans the solver failed to build. Counted as disagreements,
with the 135 other plans the solver failed to build left out as Appendix
B (decision 10) scopes them, they bring agreement on the first failing
check over the remaining 30,337 plans both ingests accept to 99.6 per
cent, the headline figure of Section 4.1. The 38 in which the ingests
disagree lie outside the 26,126 plans of Section 4.1, which both ingests
accepted; the other 57, the different-ring and quantisation classes, lie
inside it, and are the 57 plans on which the two implementations stop at
different checks there.

The 53 plans on which the two sides measure different rings are replayed
at the other side's settings one stage at a time. In 48 the deciding
room's verdict separates at the quantisation step, where each side snaps
coordinates to its own grid, a separation the triage's quantisation
test, which hands the checker geometry the solver has already quantised,
does not detect, so these plans stay in the ring class: the deciding
rooms' rings differ by a median Hausdorff distance of 0.0164 m, and the
checker's score lies within 0.005 of its threshold on 47 of the 48, so
the step is a stated difference of the two implementations rather than
an error of either. In the other 5 the two sides hand one closed seam
region, a sliver between rooms that each ingest closes into a room, to
different rooms. Figure S1 draws the three classes.

\begin{figure}
\centering
\includegraphics[width=0.75\linewidth,height=\textheight,keepaspectratio,alt={Triage of the 95 plans on which the checker and the solver disagree on the reported run, one mechanism per plan, in three classes: the two sides measure different rings (53), parting at the quantisation step (48) or at a seam handed to different rooms (5); the two ingests disagree (38); and the two quantisation steps differ (4).}]{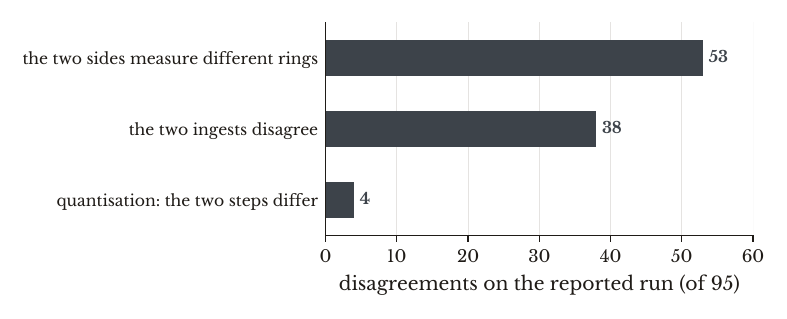}
\caption{Triage of the 95 plans on which the checker and the solver
disagree on the reported run, one mechanism per plan, in three classes:
the two sides measure different rings (53), parting at the quantisation
step (48) or at a seam handed to different rooms (5); the two ingests
disagree (38); and the two quantisation steps differ
(4).}\label{fig:disagreement_triage}
\end{figure}

\subsection{S3. Supporting
measurements}\label{s3.-supporting-measurements}

\emph{The RLVR generator.} Its four released checkpoints are
70-billion-parameter models of about 140 GB each, merged from Llama 3.3,
a release of the Llama 3 family \citep{grattafiori2024llama3}, whose
licence sampling them requires; we found no released sample set.

\emph{HouseDiffusion's overlaps.} Its plans are refused when floor is
claimed twice beyond one part per million of the floor, and 87.5 per
cent of those plans claim more than 0.1 square metres twice. A tolerance
of 0.1 square metres would still refuse 67.5 per cent of its plans. The
overlaps are therefore not rounding residue.

\emph{Training-set exposure.} House-GAN++ trains on RPLAN less one
unrecorded room count of five to eight. Taking its room count to be
RPLAN's, one of its four buckets was unseen and 75 per cent of its
graphs were seen. HouseDiffusion holds out no count, so 75 to 100 per
cent of its graphs were seen. Whichever count House-GAN++ held out, the
proportion check alone refuses the unseen bucket's admitted plans at
32.0 to 36.4 per cent, against 32.7 to 34.2 for the seen ones. GSDiff
passes 4.0 per cent of its 2,911 held-out test plans, but only 3.0 per
cent of 3,528 stride-20 RPLAN plans, 91.2 per cent of whose conditioning
graphs it saw in training. House-GAN, trained on the LIFULL HOME'S
corpus \citep{nauata2020housegan}, saw none of these plans.

\emph{Why the three margins differ.} The four terms of the graded margin
in Table 6, read from Table 2, sum to 12.3 from the rounded cells, the
12.1 net of rounding. Dropping the ingest refusals lifts House-GAN++
from 58.2 to 65.4 per cent and the MSD apartments only from 70.3 to
70.9, so the admitted margin is 6.6 points lower, at 5.5; dropping as
well the 2,879 precondition stops, all on the built side, lifts the MSD
apartments to 88.9 and the margin to 23.5.

\emph{How buildings are inferred.} The MSD release carries no building
identifier. Two floors are linked when the intersection over union of
their structural elements on a 0.5 m grid reaches one half, and a
building is a connected group of linked floors, a proxy that treats a
ground floor whose structure differs from the storeys above as a
building of its own. Resampled by building, the intervals of Section
4.2.3 are about 1.4 times as wide as plan-independent ones over graded
and over admitted plans and about 1.2 times over plans that reached a
verdict.

\emph{Tracing and orientation.} The room that decides a House-GAN++ plan
at the proportion check has a median of 145 mask pixels, and the opening
of Section 4.2.2 clears it in 2.1 per cent of the 1,133 plans the check
decides, moving the separation the check alone draws from +28.8 to +28.4
points against RPLAN and from +28.9 to +28.1 against the MSD apartments.
These baselines are measured on the checker's path, over the plans its
ingest admits and with its room measurement; the scan behind Table S2
counts every plan through the audit's ingest and measurement and gives
29.0. A room's direction is that of its minimum-area rectangle, which
can be tilted for a rectilinear room that is not a rectangle although
its walls are not; on that rectangle, with the thresholds unchanged, the
check alone refuses 5.2 rather than 4.9 per cent of the MSD apartments
and separates House-GAN++ from them by +28.7 rather than +28.9.

\emph{Resolution.} RPLAN's rooms, drawn at 0.0703125 m, were reduced to
House-GAN++'s 0.28125 m cells by centre sampling and by majority
coverage, each at 16 grid phases, and by House-GAN++'s own training
rendering at its frame, then traced as its samples are and graded on the
checker's path; Table 5 gives the separations. For the MSD apartments,
majority coverage at four phases gives 28.7 against 28.9 unchanged and
26.7 (24.9 to 28.4) against 27.9 refitted, where 28.9 (Table 5) and 27.9
are the separations from the MSD apartments at their native resolution
with the thresholds unchanged and refitted. Majority coverage leaves 29
per cent of coarse RPLAN plans untiled at ingest, but read on every plan
or on the same plans natively the share moves by at most 0.3 points.
House-GAN++'s own rendering leaves none untiled, so its 11.1 per cent
ingest refusals are compatible with a raster representation but not
forced by it, and they fall on graphs near independent of House-GAN's
equal share, 46 in both against 49 expected by chance, so neither
generator's refusals are a property of the graphs. At House-GAN's 0.5625
m cells the grid rules give the separations of Table S1.

\LTkeeptrue

\begin{longtable}[]{@{}
  >{\raggedright\arraybackslash}p{(\linewidth - 8\tabcolsep) * \real{0.3378}}
  >{\raggedright\arraybackslash}p{(\linewidth - 8\tabcolsep) * \real{0.1899}}
  >{\raggedright\arraybackslash}p{(\linewidth - 8\tabcolsep) * \real{0.1899}}
  >{\raggedright\arraybackslash}p{(\linewidth - 8\tabcolsep) * \real{0.1412}}
  >{\raggedright\arraybackslash}p{(\linewidth - 8\tabcolsep) * \real{0.1412}}@{}}
\caption{The separation the proportion check alone draws between
House-GAN and each built corpus, in points, with the built corpus
reduced to House-GAN's 0.5625 m cells; ranges in parentheses are over
grid phases. Measured natively, the separation is 27.5 points from
RPLAN, and 27.6 unchanged and 22.0 refitted from the MSD apartments; an
empty cell was not measured.}\tabularnewline
\toprule\noalign{}
\begin{minipage}[b]{\linewidth}\raggedright
Rule at 0.5625 m
\end{minipage} & \begin{minipage}[b]{\linewidth}\raggedright
From RPLAN, thresholds unchanged
\end{minipage} & \begin{minipage}[b]{\linewidth}\raggedright
From RPLAN, thresholds refitted
\end{minipage} & \begin{minipage}[b]{\linewidth}\raggedright
From the MSD apartments, thresholds unchanged
\end{minipage} & \begin{minipage}[b]{\linewidth}\raggedright
From the MSD apartments, thresholds refitted
\end{minipage} \\
\midrule\noalign{}
\endfirsthead
\toprule\noalign{}
\begin{minipage}[b]{\linewidth}\raggedright
Rule at 0.5625 m
\end{minipage} & \begin{minipage}[b]{\linewidth}\raggedright
From RPLAN, thresholds unchanged
\end{minipage} & \begin{minipage}[b]{\linewidth}\raggedright
From RPLAN, thresholds refitted
\end{minipage} & \begin{minipage}[b]{\linewidth}\raggedright
From the MSD apartments, thresholds unchanged
\end{minipage} & \begin{minipage}[b]{\linewidth}\raggedright
From the MSD apartments, thresholds refitted
\end{minipage} \\
\midrule\noalign{}
\endhead
\bottomrule\noalign{}
\endlastfoot
Centre sampling, 64 grid phases & 25.0 (24.5 to 25.5) & 21.6 (19.6 to
24.9) & & \\
Majority coverage, 64 grid phases for RPLAN, 4 for MSD & 26.0 (24.9 to
26.8) & 23.0 (19.8 to 26.2) & 25.1 & 16.0 \\
\end{longtable}

\LTkeepfalse

House-GAN's own training rendering, each room its bounding box on the 32
by 32 frame with overlaps disjoined in the exported room order, leaves
-1.8 and 2.3 points. RPLAN so rendered is refused by the check alone at
34.2 per cent, 1,626 of its 3,998 traceable plans losing a room to the
disjoin; the same boxes traced without the disjoin are refused by the
check alone at 2.0 per cent and by compactness alone at 0.3. Like
House-GAN++'s rendering, in which RPLAN is refused at 5.8 per cent, the
control stands for a generator that reproduced its training data
exactly, and in House-GAN's RPLAN is refused as often as House-GAN's own
plans; the grid rules bound the cell size alone. Two things were not
measured: whether House-GAN's emitted masks overlap as its targets do,
since the retained samples keep no masks, and whether the 0.93 rooms
each of its plans loses come from empty masks, as the sampling record
attributes them, or from the disjoin, which runs first. Nor was the
exported room order, which decides which overlapping room is carved,
checked against the order of House-GAN's conditioning graphs.

\emph{A single threshold.} The five anchors of Section 4.2.2 are four
aspect ratios, being exact geometry, and the solver's cap on a bedroom's
proportion threshold (Section 3.1). Table S2 compares them, on the
audit's path over every plan, with the thresholds fitted per room type,
which on that path refuse 5.0 per cent of RPLAN plans and separate
House-GAN++ from it by 29.0 points. Section 4.2.2 quotes the same
separation on the checker's path, 28.8 points (Table 5): the audit's
path counts all 4,000 plans of each source through its own ingest and
room measurement, 34.0 per cent of House-GAN++ refused against 5.0 of
RPLAN, where the checker's path counts the plans its ingest admits, 33.7
against 5.0.

\LTkeeptrue

\begin{longtable}[]{@{}
  >{\raggedright\arraybackslash}p{(\linewidth - 4\tabcolsep) * \real{0.7647}}
  >{\raggedright\arraybackslash}p{(\linewidth - 4\tabcolsep) * \real{0.0941}}
  >{\raggedright\arraybackslash}p{(\linewidth - 4\tabcolsep) * \real{0.1412}}@{}}
\caption{The proportion check with one threshold for every room type
against the thresholds fitted per room type, on the audit's path over
every plan, 4,000 per source.}\tabularnewline
\toprule\noalign{}
\begin{minipage}[b]{\linewidth}\raggedright
Proportion threshold
\end{minipage} & \begin{minipage}[b]{\linewidth}\raggedright
Real RPLAN plans refused (\%)
\end{minipage} & \begin{minipage}[b]{\linewidth}\raggedright
Separation of House-GAN++ from RPLAN (points)
\end{minipage} \\
\midrule\noalign{}
\endfirsthead
\toprule\noalign{}
\begin{minipage}[b]{\linewidth}\raggedright
Proportion threshold
\end{minipage} & \begin{minipage}[b]{\linewidth}\raggedright
Real RPLAN plans refused (\%)
\end{minipage} & \begin{minipage}[b]{\linewidth}\raggedright
Separation of House-GAN++ from RPLAN (points)
\end{minipage} \\
\midrule\noalign{}
\endhead
\bottomrule\noalign{}
\endlastfoot
Fitted per room type, the reported run & 5.0 & +29.0 \\
Single, 1:3 & 28.7 & +7.7 \\
Single, 1:2.5 & 50.5 & +9.0 \\
Single, 1:2 & 79.7 & -0.5 \\
Single, the bedroom cap of 0.60 & 95.0 & -2.0 \\
Single, 1:1.5 & 98.4 & -1.6 \\
Single, held to the fitted rate, a worst-room squareness of 0.209 & held
to the fitted rate & +1.6 \\
\end{longtable}

\LTkeepfalse

\emph{The building-code rule.} IRC R304.2's 7 feet in any horizontal
dimension was applied to the finished plan under two measures of a
room's least dimension, the shorter side of its minimum-area rectangle
and the diameter of its largest inscribed circle, alone, in place of the
proportion threshold behind compactness, and behind checks 1 and 4 with
neither fitted threshold; Table S3 gives the margins, computed on the
checker's verdicts. Alone, the pair that binds changes because the
ingest refuses 70 per cent of HouseDiffusion before any room is measured
and the rule passes most of what it admits.

\LTkeeptrue

\begin{longtable}[]{@{}
  >{\raggedright\arraybackslash}p{(\linewidth - 6\tabcolsep) * \real{0.4952}}
  >{\raggedright\arraybackslash}p{(\linewidth - 6\tabcolsep) * \real{0.1627}}
  >{\raggedright\arraybackslash}p{(\linewidth - 6\tabcolsep) * \real{0.1594}}
  >{\raggedright\arraybackslash}p{(\linewidth - 6\tabcolsep) * \real{0.1827}}@{}}
\caption{The building-code width rule on the public checker's verdicts:
margins in points, the MSD apartments over House-GAN++ unless a cell
names another pair, with intervals in parentheses; a range for the width
rule alone spans the two measures of a room's least dimension, and a
range under a moved scale or threshold spans those
moves.}\tabularnewline
\toprule\noalign{}
\begin{minipage}[b]{\linewidth}\raggedright
Ladder
\end{minipage} & \begin{minipage}[b]{\linewidth}\raggedright
Graded
\end{minipage} & \begin{minipage}[b]{\linewidth}\raggedright
Admitted
\end{minipage} & \begin{minipage}[b]{\linewidth}\raggedright
Reached a verdict
\end{minipage} \\
\midrule\noalign{}
\endfirsthead
\toprule\noalign{}
\begin{minipage}[b]{\linewidth}\raggedright
Ladder
\end{minipage} & \begin{minipage}[b]{\linewidth}\raggedright
Graded
\end{minipage} & \begin{minipage}[b]{\linewidth}\raggedright
Admitted
\end{minipage} & \begin{minipage}[b]{\linewidth}\raggedright
Reached a verdict
\end{minipage} \\
\midrule\noalign{}
\endhead
\bottomrule\noalign{}
\endlastfoot
The checker's fitted ladder & +12.2 & +5.6 & +23.6 \\
Width rule alone & +22.3, RPLAN over House-GAN++ & +6.6 to +11.6, RPLAN
over HouseDiffusion & \\
Width rule in place of the proportion threshold & +1.7 (+0.1 to +3.3) &
-6.6 (-8.1 to -5.0) & +10.6 (+9.1 to +12.3), RPLAN over House-GAN++ \\
Same, scales moved 5 per cent either way or the width threshold one
frame step & -4.8 to +8.7 & -13.9 to +1.3 & +3.3 to +18.5 \\
Checks 1 and 4, then the width rule & +3.0 & -5.5 & +2.9, RPLAN over
HouseDiffusion \\
Same, scales moved 5 per cent either way or the width threshold one
frame step & -3.8 to +10.3 & & \\
\end{longtable}

\LTkeepfalse

The output frames of House-GAN++, House-GAN and HouseDiffusion carry no
metre factor and are read over RPLAN's 18 m, at 0.28125 m per unit for
House-GAN++, 0.5625 m for House-GAN and 0.0703125 m for HouseDiffusion.
For the substituted rule, the low end of each range in Table S3 comes
with the width threshold one step lower and the high end with it one
step higher. As exported and at the larger scale the substituted rule
recovers less than half of the fitted ladder's margin on every
denominator and reverses it over admitted plans; at the smaller scale it
recovers just over half over plans that reached a verdict, 12.4 of 23.6
points, and with the width threshold one step higher it recovers 8.7 of
12.2 over graded plans and 18.5 of 23.6 over plans that reached a
verdict, so the comparison holds at the inferred scale and threshold and
not across the span.

\emph{How Compatibility was computed.} It is the graph edit distance
between the requested room-adjacency graph and the graph of rooms that
share at least a door's width of wall in the generated plan, with room
types as node labels: two rooms are adjacent when at least 0.813 m of
one room's boundary, a door's width, lies within 0.3 m of the other's.
The generators' own evaluators recover and score the output graph
differently from this rule and from each other, from bounding boxes with
room labels unmatched, from masks or polygons at their own resolution
under a fixed room correspondence, or from identical shared edges, so
the absolute values here are not comparable with published ones:
HouseDiffusion's score on the graph of rooms joined by a drawn door, its
own evaluator's convention, is 1.58 on five-room and 3.15 on eight-room
tasks, near the 1.5 and above the 2.5 that \citet{lara2026rlvr} report
for the strongest prior diffusion baseline, and GSDiff's 1.25 lies above
its own published value of about 0.49. The claims of Section 5.1
therefore rest on comparisons within one generator and on the order of
the two diffusion generators against the two GANs, not on the absolute
values. The House-GAN++ plans first rejected at the proportion check
match their request slightly better than those that pass, a mean
distance of 3.87 against 4.09 House-GAN and House-GAN++ produce 4,275 of
the 4,434 generated plans the ladder passes, and a plan that matches its
request exactly passes at 55 to 65 per cent, about its generator's own
rate. The diffusion generators' passing plans number 42 for
HouseDiffusion and 117 for GSDiff.

\emph{Intervals on the AUCs.} Table S4 gives each generator's AUC for
predicting a pass with a 95 per cent percentile interval from 2,000
bootstrap resamples of its plans, and the counts behind it. Within
House-GAN and House-GAN++, each cell rests on 1,948 or 2,327 passing
plans against 1,230 to 2,052 failing ones. Within House-GAN and
House-GAN++ every point estimate lies within 0.07 of 0.5; where an
interval excludes 0.5 it does so on either side, so the relation is weak
and inconsistent in sign rather than exactly nil. The diffusion
generators' few passing plans give intervals two to five times as wide.

\LTkeeptrue

\begin{longtable}[]{@{}
  >{\raggedright\arraybackslash}p{(\linewidth - 10\tabcolsep) * \real{0.1765}}
  >{\raggedright\arraybackslash}p{(\linewidth - 10\tabcolsep) * \real{0.1294}}
  >{\raggedright\arraybackslash}p{(\linewidth - 10\tabcolsep) * \real{0.0941}}
  >{\raggedright\arraybackslash}p{(\linewidth - 10\tabcolsep) * \real{0.0941}}
  >{\raggedright\arraybackslash}p{(\linewidth - 10\tabcolsep) * \real{0.2529}}
  >{\raggedright\arraybackslash}p{(\linewidth - 10\tabcolsep) * \real{0.2529}}@{}}
\caption{The area under the ROC curve of Compatibility for predicting a
pass, within each generator, over graded (G) and admitted (A) plans,
pooled and over pairs within one requested room count, with 95 per cent
bootstrap intervals; 0.5 is no relation.}\tabularnewline
\toprule\noalign{}
\begin{minipage}[b]{\linewidth}\raggedright
Generator
\end{minipage} & \begin{minipage}[b]{\linewidth}\raggedright
Population
\end{minipage} & \begin{minipage}[b]{\linewidth}\raggedright
Passing
\end{minipage} & \begin{minipage}[b]{\linewidth}\raggedright
Failing
\end{minipage} & \begin{minipage}[b]{\linewidth}\raggedright
AUC, pooled
\end{minipage} & \begin{minipage}[b]{\linewidth}\raggedright
AUC, within bucket
\end{minipage} \\
\midrule\noalign{}
\endfirsthead
\toprule\noalign{}
\begin{minipage}[b]{\linewidth}\raggedright
Generator
\end{minipage} & \begin{minipage}[b]{\linewidth}\raggedright
Population
\end{minipage} & \begin{minipage}[b]{\linewidth}\raggedright
Passing
\end{minipage} & \begin{minipage}[b]{\linewidth}\raggedright
Failing
\end{minipage} & \begin{minipage}[b]{\linewidth}\raggedright
AUC, pooled
\end{minipage} & \begin{minipage}[b]{\linewidth}\raggedright
AUC, within bucket
\end{minipage} \\
\midrule\noalign{}
\endhead
\bottomrule\noalign{}
\endlastfoot
House-GAN & G & 1,948 & 2,052 & 0.504 (0.487 to 0.521) & 0.451 (0.433 to
0.467) \\
House-GAN & A & 1,948 & 1,607 & 0.487 (0.468 to 0.506) & 0.432 (0.413 to
0.452) \\
House-GAN++ & G & 2,327 & 1,673 & 0.531 (0.513 to 0.548) & 0.519 (0.501
to 0.536) \\
House-GAN++ & A & 2,327 & 1,230 & 0.472 (0.453 to 0.490) & 0.477 (0.458
to 0.496) \\
HouseDiffusion & G & 42 & 3,949 & 0.777 (0.724 to 0.826) & 0.666 (0.599
to 0.729) \\
HouseDiffusion & A & 42 & 982 & 0.645 (0.574 to 0.712) & 0.607 (0.532 to
0.681) \\
GSDiff & G & 117 & 2,794 & 0.579 (0.539 to 0.616) & 0.540 (0.494 to
0.586) \\
GSDiff & A & 117 & 2,684 & 0.563 (0.522 to 0.601) & 0.525 (0.475 to
0.571) \\
\end{longtable}

\LTkeepfalse

\subsection{S4. Room labels and threshold
types}\label{s4.-room-labels-and-threshold-types}

The proportion check holds each room to the threshold of its room type
(Table 1). Table S5 gives how each source's room labels reach those
types. A label is lower-cased and looked up in the checker's stated
table; a label the table does not state becomes Unset, which takes no
proportion threshold, so the proportion check skips the room, while the
room still counts in the ingest's tiling and in the compactness check,
and the plan is neither refused nor reduced. Generated plans are graded
on the RPLAN thresholds, which exist for every type in the table but
Hallway. An unmapped label can therefore only lower a source's refusal
at the proportion check. Recorded on the reported run, 548 House-GAN++
study rooms, 80 House-GAN closets and 49 dining rooms, and 2,918 GSDiff
living, dining and foyer rooms fall to Unset.

\LTkeeptrue

\begin{longtable}[]{@{}
  >{\raggedright\arraybackslash}p{(\linewidth - 4\tabcolsep) * \real{0.1529}}
  >{\raggedright\arraybackslash}p{(\linewidth - 4\tabcolsep) * \real{0.5705}}
  >{\raggedright\arraybackslash}p{(\linewidth - 4\tabcolsep) * \real{0.2765}}@{}}
\caption{How each source's room labels reach the threshold room types; a
label named without a type reaches the type of the same
name.}\tabularnewline
\toprule\noalign{}
\begin{minipage}[b]{\linewidth}\raggedright
Source
\end{minipage} & \begin{minipage}[b]{\linewidth}\raggedright
Labels and the type each reaches
\end{minipage} & \begin{minipage}[b]{\linewidth}\raggedright
Labels that fall to Unset
\end{minipage} \\
\midrule\noalign{}
\endfirsthead
\toprule\noalign{}
\begin{minipage}[b]{\linewidth}\raggedright
Source
\end{minipage} & \begin{minipage}[b]{\linewidth}\raggedright
Labels and the type each reaches
\end{minipage} & \begin{minipage}[b]{\linewidth}\raggedright
Labels that fall to Unset
\end{minipage} \\
\midrule\noalign{}
\endhead
\bottomrule\noalign{}
\endlastfoot
RPLAN & livingroom, LivingRoom; masterroom, PrimaryBedroom; childroom
and secondroom, Bedroom; guestroom, GuestRoom; kitchen; bathroom;
diningroom, DiningRoom; balcony; entrance, Foyer; storage and
walkin\_closet, Closet & studyroom \\
MSD & bedroom; livingroom, LivingRoom; kitchen; dining, DiningRoom;
corridor, Hallway; storeroom, Closet; bathroom; balcony & stairs, any
unknown code \\
House-GAN++, HouseDiffusion & living\_room, LivingRoom; bedroom, from
RPLAN's primary, child, second and guest bedrooms, Bedroom; kitchen;
bathroom; balcony; entrance, Foyer; dining room, DiningRoom; storage,
from RPLAN's storage and walk-in closet, Closet & study room, unknown \\
House-GAN & living\_room, LivingRoom; bedroom, from RPLAN's bedrooms and
study room, Bedroom; kitchen; bathroom; balcony; corridor, from RPLAN's
entrance, Hallway & closet, dining\_room, missing, laundry\_room \\
GSDiff & bedroom; storage, Closet; kitchen; bathroom; balcony &
living/dining/foyer; exterior is excluded before export \\
\end{longtable}

\LTkeepfalse

Three asymmetries follow from the label sets. House-GAN++ and
HouseDiffusion have no primary or guest bedroom, so such a room is held
to Bedroom, 0.4747, rather than 0.4553 or 0.4655; on House-GAN++ this
accounts for at most 1.4 points of its refusal (Section 4.2.2).
House-GAN's closets and dining rooms take no threshold because of their
spelling, and its corridor, RPLAN's entrance, becomes Hallway, which has
no RPLAN threshold, where RPLAN's entrance is a Foyer held to 0.3376.
GSDiff's merged living, dining and foyer class takes no threshold. The
last two lighten the check on House-GAN and GSDiff. GSDiff's was
recomputed with the merged class held to the RPLAN threshold of
LivingRoom, 0.3486, which is also the strictest of LivingRoom,
DiningRoom, 0.3098, and Foyer, 0.3376: the proportion check alone then
refuses 7.7 per cent of its 2,820 admitted plans rather than 6.9,
against 5.0 for RPLAN, and the ladder passes the same 117 plans.
House-GAN's was not recomputed.

\end{document}

%% file: layout.tex
\usepackage{booktabs}
\usepackage{longtable}
\usepackage{etoolbox}
\ifdefined\venuestrictformat
  \AtBeginEnvironment{longtable}{\setlength{\tabcolsep}{4pt}}
  \makeatletter
  \@ifpackageloaded{caption}{\captionsetup[table]{skip=\abovecaptionskip}}{}
  \makeatother
\else
\fi

\newif\ifLTkeep
\makeatletter
\protected\def\LT@keep@tabularcr{%
  \relax\ifnum0=`}\fi
  \def\crcr{\LT@crcr\noalign{\nobreak}}\let\cr\crcr
  \LT@t@bularcr}
\AtBeginEnvironment{longtable}{%
  \footnotesize
  \ifLTkeep\let\LT@tabularcr\LT@keep@tabularcr\fi}
\ExplSyntaxOn
\patchcmd{\LT@start}{\global\@colroom\@colht}{}{}
  {\PackageError{layout}{Could~not~patch~longtable's~\string\LT@start}{}}
\def\LT@newpage@room{
  \global\@colroom\@colht
  \ifvoid\LT@foot\else\global\advance\@colroom-\ht\LT@foot\fi
  \global\vsize\@colroom}
\tl_set:No \l_tmpa_tl { \LT@output }
\tl_if_in:NnF \l_tmpa_tl { \global\vsize\@colroom }
  { \PackageError{layout}{Could~not~patch~longtable's~\string\LT@output}{} }
\tl_replace_all:Nnn \l_tmpa_tl { \global\vsize\@colroom } { \LT@newpage@room }
\exp_args:NNV \cs_set:Npn \LT@output \l_tmpa_tl
\ExplSyntaxOff
\makeatother

\usepackage{float}
\ifdefined\venuestrictformat\else

\fi